\documentclass{article} 
\usepackage{iclr2026_conference,times}

\usepackage{amsmath}
\usepackage{booktabs}
\usepackage{float}
\usepackage[T1]{fontenc}
\usepackage{graphicx}
\usepackage{hyperref}
\usepackage{multirow}
\usepackage{url}

\title{MolPILE - large-scale, diverse dataset for molecular representation learning}

\renewcommand*{\thefootnote}{\fnsymbol{footnote}}

\author{\vspace{-2mm}
    \textbf{Jakub Adamczyk}$^{1,*}$, 
    \textbf{Jakub Poziemski}$^{2}$,
    \textbf{Franciszek Job}$^{1}$, 
    \textbf{Mateusz Król}$^{1}$, 
    \textbf{Maciej Makowski}$^{1}$  \\ \\
    $^{1}$Faculty of Computer Science, AGH University of Krakow, Cracow, Poland \\
    $^{2}$Institute of Biochemistry and Biophysics, Polish Academy of Sciences, Warsaw, Poland \\
    $^*$Corresponding author, \texttt{jadamczy@agh.edu.pl}, ORCID: 0000-0003-4336-4288
    \vspace{-5mm}
}

\def\customfootnotetext#1#2{{%
  \let\thefootnote\relax
  \footnotetext[#1]{#2}}}

\iclrfinalcopy 
\begin{document}

\maketitle

\begin{abstract}
The size, diversity, and quality of pretraining datasets critically determine the generalization ability of foundation models. Despite their growing importance in chemoinformatics, the effectiveness of molecular representation learning has been hindered by limitations in existing  small molecule datasets. To address this gap, we present MolPILE, large-scale, diverse, and rigorously curated collection of 222 million compounds, constructed from 6 large-scale databases using an automated curation pipeline. We present a comprehensive analysis of current pretraining datasets, highlighting considerable shortcomings for training ML models, and demonstrate how retraining existing models on MolPILE yields improvements in generalization performance. This work provides a standardized resource for model training, addressing the pressing need for an ImageNet-like dataset in molecular chemistry.
\end{abstract}

\section{Introduction}


Modern chemoinformatics relies extensively on machine learning (ML) methods, particularly for virtual screening and ADMET workflows. In practice, however, drug design problems are often constrained to very small datasets, typically comprising only hundreds or a few thousand molecules \citep{low_data_learning}. This presents a major limitation for traditional workflows based on feature extraction methods, such as molecular fingerprints \citep{scikit-fingerprints}, which are subsequently paired with tabular classifiers. Molecular representation learning has emerged as a powerful strategy for incorporating general chemical knowledge from large-scale molecular databases. The transfer learning capabilities of pretrained neural networks mitigate data scarcity either through fine-tuning for specific targets or by employing embeddings from these models as pretrained feature extractors \citep{pretrained_feature_extractors}. The latter approach is particularly important for unsupervised tasks common in chemoinformatics, including molecular clustering \citep{molecular_clustering} and similarity searching \citep{molecular_similarity_searching}.

The amount, diversity, and quality of the pretraining dataset directly influence the transfer learning capabilities of models \citep{data_influence,data_influence_2,data_influence_3}. This is particularly critical in chemistry, which encompasses a wide range of subdomains and applications, extending beyond medicinal chemistry (itself defined by numerous ADMET properties) to, e.g., natural products \citep{natural_product_score}, agrochemistry and ecotoxicology \citep{ApisTox}, materials science and industrial chemistry \citep{materials_science}, or food and flavor chemistry \citep{flavor_chemistry}. These domains differ substantially in typical structures, functional groups, and element distributions. Pretrained models must therefore perform robustly not only for organic molecules but also for metals, organometallics, salts, and other classes relevant to established QSAR workflows \citep{QSAR_workflows}. Equally important is data quality: pretraining sets should reflect realistic experimental conditions through rigorous deduplication, standardized chemical structures, molecular weight ranges appropriate for small molecules (excluding, e.g., proteins, peptides, or nucleic acids), and constraints on properties such as synthetic accessibility and solubility (e.g., avoiding compounds with extreme logP values).

Deficiencies in existing datasets in the aforementioned aspects have significant implications for ML models. A recent study of \cite{pretrained_feature_extractors} reported that nearly all existing pretrained neural models perform statistically worse than the simple ECFP molecular fingerprint \citep{ECFP} on a large-scale benchmark. Notably, the majority of these models were pretrained on very limited subsets of ZINC or PubChem, without assuring diversity or quality filtering (see Appendix \ref{appendix_model_pretraining_datasets} for a detailed overview), which may represent a key factor underlying this outcome. Similarly, in \cite{data_influence_chemistry} authors point out the importance of chemical space coverage and data filtering for pretraining molecular models, and note that the disparity in pretraining datasets makes it considerably harder to fairly compare the generalization of models. Their performance differences may not stem from algorithmic advances, but rather pretraining data used. This situation shows the need for a single, high-quality dataset for pretraining molecular ML models.

\textbf{Key contributions} of this work are as follows. Primarily, we introduce the \textbf{MolPILE dataset}, a large-scale collection of small compounds for molecular representation learning and neural model pretraining, designed to satisfy the key desiderata of size, diversity, and quality. With 222 million molecules, it is the largest publicly available dataset of experimentally verified, synthesizable compounds. It spans a broad chemical space, encompassing substantial variation in elements, structures, and properties. A multi-step processing and filtering workflow further ensures data quality through deduplication, structure standardization, and real-world feasibility filtering. In essence, it is intended to serve a role for molecular machine learning comparable to that of ImageNet in computer vision and PILE in natural language processing, providing a unified and standardized pretraining dataset.

In addition, we provide a \textbf{comprehensive examination of the chemical spaces} represented in commonly used pretraining datasets, including UniChem \citep{UniChem}, PubChem \citep{PubChem}, and ZINC \citep{ZINC}. Comparative analyses demonstrate the superior diversity and quality of MolPILE, while also identifying specific deficiencies in existing datasets.

Finally, we show how these dataset characteristics \textbf{translate directly into model performance}. Models such as Mol2vec \citep{Mol2vec} and ChemBERTa \citep{ChemBERTa}, when trained on MolPILE, achieve results surpassing those of their original counterparts.

Dataset is available at \href{https://huggingface.co/datasets/scikit-fingerprints/MolPILE}{https://huggingface.co/datasets/scikit-fingerprints/MolPILE}, and code at \href{https://github.com/scikit-fingerprints/MolPILE_dataset}{https://github.com/scikit-fingerprints/MolPILE\_dataset}.

\section{Literature review}

\textbf{Molecular databases creation.} Large-scale small molecule collections can be constructed primarily in three ways: experimentally, by combinatorial enumeration, or with combinatorial reactions.

\textit{Experimental databases} are the most established resources, containing molecules that have been synthesized and tested for specific purposes, such as high-throughput screening (HTS). Their data typically originates from scientific literature, patents, and commercial suppliers. Large aggregate repositories include PubChem \citep{PubChem}, UniChem \citep{UniChem}, and ZINC15 \citep{ZINC}, and there also exist more specialized examples, such as COCONUT \citep{COCONUT} and SuperNatural3 \citep{SuperNatural3} for natural products. These collections are generally diverse and realistic, reflecting contributions from a wide range of projects and subfields in chemistry. At present, experimental databases encompass at most roughly 200 million small molecules, though data quality can be inconsistent. For instance, vendors sometimes submit larger biomolecules such as peptides or RNAs, or molecules not parseable by RDKit.

\textit{Combinatorially enumerated} databases are constructed by generating all theoretically possible atomic structures and bonds up to a given size, followed by filtering out chemically implausible arrangements based on predefined rules. A prominent example is GDB-17 \citep{GDB17}. While such databases can reach immense sizes, their diversity, quality, and synthesizability are constrained by the rules applied during generation and filtering, which are difficult to enforce consistently across the entire dataset \citep{chemical_databases_4}. Despite their scale, these databases often cover only limited regions of chemical space in terms of elements and scaffolds, and are restricted to relatively small molecules due to combinatorial explosion at higher atom counts.

\textit{Combinatorial reaction} databases generate compounds by combining readily available reagents using well-defined chemical reactions, often described with representations such as SMARTS or LHASA patterns. Examples include ZINC20 \citep{ZINC20} and Enamine REAL \citep{EnamineREAL}. This approach enables the creation of large, synthetically accessible datasets by leveraging well-established reaction pathways. However, the resulting chemical space is inherently constrained, as it primarily reflects common and safe routes for synthesis. Moreover, the quality and diversity of the generated compounds depend heavily on the availability and quality of the building blocks used.

\textbf{Database filtering.} Many commercial molecular databases apply extensive filtering, such as Lipinski’s rule of 5 \citep{Lipinski}, which has been criticized for biasing chemical space toward narrow physicochemical ranges \citep{Lipinski_critique_1,Lipinski_critique_2,Lipinski_critique_3}. For instance, Enamine REAL \citep{EnamineREAL} applies Lipinski and Veber filters \citep{filters_Veber}, Maybridge enforces a variety of structural rules and almost entirely conforms to Lipinski rule of 5 \citep{Maybridge}, and ChemDiv uses REOS \citep{REOS} and PAINS \citep{PAINS} filters \citep{ChemDiv}. While effective for targeted screening, such preprocessing substantially reduces chemical diversity and limits usefulness for ML pretraining. Broader and more representative coverage is provided by unfiltered sources such as ChemSpace \citep{ChemSpace}, Mcule \citep{Mcule}, PubChem, and UniChem.

\textbf{Pretraining of molecular ML models.} Foundation models benefit from larger parameter counts and datasets, following neural scaling laws \citep{neural_scaling_laws}, but their performance also depends on the quality and diversity of training data \citep{neural_scaling_laws_quality,data_influence,data_influence_chemistry}. In molecular ML, pretraining datasets are often limited to small subsets of ZINC15 or ChEMBL, typically under 20 million compounds, which cannot fully represent real-world chemical diversity. Examples include SimSon \citep{SimSon} (1M) and R-MAT \citep{pretrained_RMAT} (4M). At the other extreme, some models train on massive but unfiltered collections, such as ChemFM \citep{ChemFM} (entire UniChem) and ChemBERTa \citep{ChemBERTa} (77M subset of PubChem), which include abnormal or non-synthesizable compounds. A third line of work uses subsets of prefiltered databases, or applies additional filtering criteria, which further constrains the available chemical space, as in GEM \citep{GEM} (20M from ZINC) or CDDD \citep{CDDD} (six custom filters on ZINC and PubChem). Appendix \ref{appendix_model_pretraining_datasets} provides a detailed comparison.

\section{MolPILE dataset}

\begin{figure}
    \centering
    \includegraphics[width=0.8\textwidth]{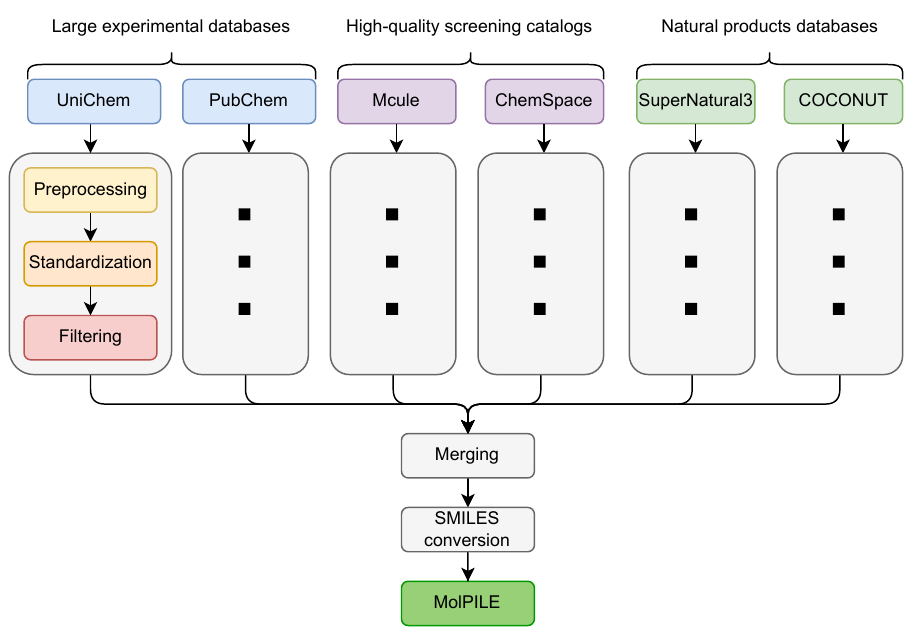}
    \caption{MolPILE data processing workflow.}
    \label{workflow}
    \vspace{-3mm}
\end{figure}

The main contribution of this work is MolPILE, a large-scale dataset of nearly 222 million small compounds for pretraining molecular ML models. It is created based on multiple sources, with workflow illustrated in Figure \ref{workflow}, with focus on quality and diversity. Each source database was processed using a uniform three-stage pipeline consisting of preprocessing, standardization, and filtering. Resulting datasets were merged and finally deduplicated, yielding the MolPILE dataset.

\subsection{Source data}

The foundation of MolPILE is built on UniChem and PubChem, two large general-purpose chemical databases comprising experimentally synthesized compounds. These databases aggregate data from multiple suppliers and other databases; for example, UniChem includes the entirety of ChEMBL \citep{ChEMBL}, providing MolPILE with broad coverage of chemical sources. UniChem applies stringent input filtering rules based on the InChI format \citep{InChI}, whereas PubChem accepts a wider range of input data and vendors. See Appendix \ref{appendix_unichem_pubchem_sources} for an overview of popular sources included in these databases.

We also incorporate Mcule \citep{Mcule} and ChemSpace \citep{ChemSpace}, screening catalogs containing commercially available molecules. As such, they are by design high-quality, readily synthesizable, and highly diverse. We selected these two sources because, unlike many alternatives, they do not apply restrictive filters such as the Lipinski Rule of 5 \citep{Lipinski}. Aggressive filtering can negatively impact the generalization of chemical foundation models, as it drastically limits chemical space coverage, a point widely discussed in drug design literature \citep{Lipinski_critique_1,Lipinski_critique_2,Lipinski_critique_3}.

Finally, we include SuperNatural3 \citep{SuperNatural3} and COCONUT \citep{COCONUT}, two large natural product (NP) databases. Interest in NP-based therapeutics is growing due to their potent bioactivity and recent successes in the field, such as Paclitaxel \citep{natural_products_review}. However, NPs are underrepresented in typical databases, which primarily focus on synthetic medicinal compounds. Incorporating these NP datasets enable MolPILE to more densely sample NP-like chemical space and enhances their representation in pretrained models.

We select only experimental databases without additional filtering. This ensures their diversity and avoids biasing MolPILE towards any specific subset of chemical compounds.

\subsection{Processing workflow}

The processing workflow is divided into three distinct phases, described below. Each phase removes molecules that cannot be parsed at that step and performs deduplication based on the InChI format. InChI was chosen as the primary representation because multiple studies have demonstrated its advantages in structure canonicalization and unambiguous representation, which are crucial for data deduplication and minimizing structural redundancy \citep{InChI_advantages_1,InChI_advantages_2,InChI_advantages_3}. RDKit is employed as the main molecular processing framework, offering an open-source environment. See Appendix \ref{appendix_implementation_and_hardware} for implementation and hardware details.

In the \textbf{preprocessing} step, each dataset is converted into a Parquet file containing molecules in InChI format. The diversity of input formats and representations presents a major challenge when integrating data from multiple sources. For example, UniChem provides a CSV with InChI strings, whereas ChemSpace is distributed as an SDF file. After preprocessing, all datasets share a consistent format, allowing subsequent steps to be applied uniformly across datasets.

The \textbf{standardization} step focuses on unifying molecular graph representations. They can vary across datasets depending on the tools and frameworks used, leading to subtle structural duplicates and differences, for example in functional group representation. Such inconsistencies can introduce noise during model pretraining. To address this, we implement standard sanitization and cleanup procedures in RDKit, as recommended for general QSAR workflows \citep{QSAR_workflows}, including kekulization, valence checks, aromaticity modeling, conjugation and hybridization adjustments, hydrogen removal, metal disconnection, functional group normalization, and reionization. Molecules that fail any of these steps are removed. We intentionally avoid aggressive filtering, such as salt removal, since it is not recommended in certain domains, e.g., agrochemistry \citep{ApisTox}.

Finally, we introduce a novel \textbf{molecular feasibility filter} as part of the filtering step. Its purpose is to remove erroneous molecules, such as those with physically unreasonable values of properties like logP or TPSA, multi-fragment complexes, and excessively large compounds, including potential peptides, crystals, or metal clusters. To design the filter, we analyzed the distributions of these properties in standardized datasets and relevant literature (see Appendix \ref{appendix_feasibility_filter_design} for details). The filter criteria are as follows: molecular fragments $\leq 3$, length of InChI $< 2000$, molecular weight $\leq 2500$, number of atoms $\leq 150$, HBA $\leq 20$, HBD $\leq 15$, logP in range $[-10, 25]$, TPSA $\leq 500$, number of rotatable bonds $\leq 60$. Those conditions are intentionally lax, aiming only to remove clearly infeasible molecules without unduly restricting chemical space. This feasibility filter can also serve as a general data quality tool for other molecular ML workflows.

\subsection{Merging and subset selection}

After processing all datasets, we merge them. During all steps, we keep a unique ID for each molecule, e.g. CID in PubChem. In the merge step, we incorporate the datasets sequentially, adding molecules from each source in order of decreasing dataset size. This approach preserves full traceability, allowing us to unambiguously link every molecule to its original database, and also ensures no duplicates. Appendix \ref{appendix_source_datasets_contributions} provides a detailed breakdown of inter-dataset contributions.

Lastly, we convert all molecules from InChI to SMILES format, as SMILES is the representation most widely adopted in chemical language models and is readily parsed by standard chemoinformatics libraries. The final MolPILE dataset comprises nearly \textbf{222 million molecules}. It is freely and publicly available; see Appendix \ref{appendix_licensing} for exact licensing.

To facilitate experimentation and the training of computationally intensive models, we additionally provide diverse subsets containing 1M, 5M, and 10M molecules. These subsets are constructed using an algorithm inspired by diversity-based selection in virtual screening. Full algorithmic details are provided in Appendix \ref{appendix_diversity_subsets}; here we present a brief overview. We employ the MaxMin approximation to maximum diversity picking \citep{MaxMin_RDKit,maximum_diversity_picking}, which identifies a subset of $k$ molecules that maximizes the sum of their pairwise Tanimoto distances. The dataset is then clustered by assigning each compound to its nearest selected molecule, as measured by Tanimoto distance. Molecules are subsequently sampled uniformly from each cluster until the desired subset size is obtained.

\section{Datasets analysis}

In this section, we perform a series of analyses for evaluating the size, diversity, and quality of MolPILE and its source datasets. We also compare with alternative datasets, commonly used for pretraining molecular representation models, with results indicating particular deficiencies in each one in at least one of those three areas.

ChEMBL \citep{ChEMBL} contains bioactivity data from screening assays. While it is included in MolPILE through UniChem, we analyze it separately given its frequent use in model pretraining. GDB-17 \citep{GDB17} is an enumerated dataset comprising compounds with up to 17 atoms of $C$, $N$, $O$, $S$, and halogens; we employ the publicly available representative subset of 50 million molecules. ZINC15 \citep{ZINC} provides experimentally validated small molecules with an emphasis on medicinal chemistry and rigorous filtering. We use its high-quality subset of 13.7 million molecules, with established 3D structures and verified in-stock vendor availability. We choose ZINC15 rather than ZINC20, as it is more commonly used in model pretraining, and the latter is also almost exclusively combinatorial (enumerated).


\subsection{Dataset size}

\begin{table}
\centering
\caption{Dataset sizes: initial count, molecules removed at each step, and the final count.}
\resizebox{\textwidth}{!}{
\begin{tabular}{lrrrrr}
\toprule
\textbf{Dataset} & 
\multicolumn{1}{r}{\textbf{Initial count}} & 
\multicolumn{1}{r}{\textbf{Preprocessing}} & 
\multicolumn{1}{r}{\textbf{Standardization}} & 
\multicolumn{1}{r}{\textbf{Filtering}} & 
\multicolumn{1}{r}{\textbf{Final dataset}} \\
\midrule
UniChem       & 189M & 0        & -459k  & -4.7M & 184M \\
PubChem       & 121.4M & -467k  & -107k  & -4.2M & 116.6M \\
Mcule         & 43.6M  & -105k  & -157     & -13.6k   & 43.4M  \\
ChemSpace     & 7.8M   & -78      & -107     & -4.4k    & 7.8M   \\
SuperNatural3 & 1.2M   & -2.9k    & -331     & -44k   & 1.1M   \\
COCONUT       & 695k    & -8.6k    & -154     & -26k   & 660k    \\
\hline
\addlinespace[2pt] 
ChEMBL        & 2.4M  & -8      & -911   & -41k  & 2.4M \\
GDB-17        & 50M & -4.3k  & -5     & 0        & 50M \\
ZINC          & 13.7M & -945k & -8.7k & -3k   & 12.7M \\
\hline
\addlinespace[2pt] 
\textbf{MolPILE} & \multicolumn{4}{c}{} & \makebox[0pt][r]{222M (221,950,487)} \\
\bottomrule
\end{tabular}
}
\label{dataset_workflow_statistics}
\end{table}

The final MolPILE dataset contains nearly 222 million molecules. To illustrate the impact of our multi-step workflow, Table \ref{dataset_workflow_statistics} summarizes the number of compounds removed at each step (rounded for readability; see Appendix \ref{appendix_pipeline_filtering_details} for exact numbers). MolPILE has about 33M more compounds than its largest source, UniChem. Overall, the filtering step removes the largest fraction of molecules, particularly from the UniChem and PubChem datasets. This is expected, given the large scale and heterogeneous sources of these databases, which include many lower-quality molecules. In total, our processing pipeline removes over 10M molecules. For ChEMBL and GDB-17, our pipeline would remove almost no compounds, indicating their good quality. However, the main problem with ChEMBL is its small size; at just 2.4M molecules, it is not suitable for training any large-scale models, particularly transformer-based. ZINC loses about 1M molecules out of initial 13.7M, which is quite concerning. These results highlight the importance of high-quality, curated datasets like MolPILE, as almost all models are currently pretrained on raw data containing such erroneous structures.

\subsection{Element groups}

\begin{figure}
    \centering
    \includegraphics[width=0.75\textwidth]{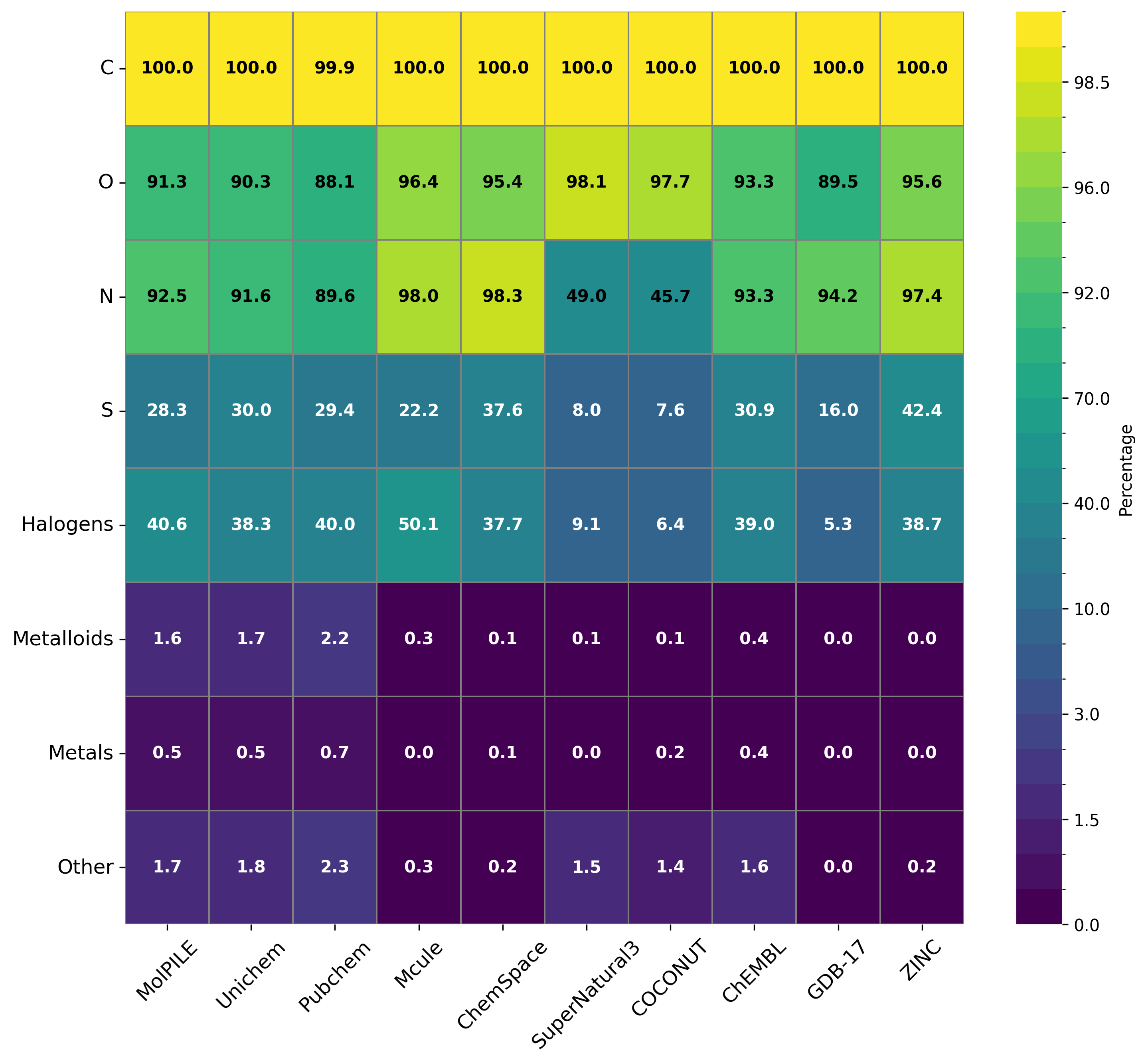}
    \caption{Percentage of molecules containing particular element or elements group.}
    \label{atom_groups}
\end{figure}

The presence and frequency of different atomic elements, such as $C$, $O$, $N$, $S$, as well as halogens, metalloids, and metals, directly influence model capabilities. Pretraining on data lacking certain elements typically groups them as an ``other'' type, which can substantially reduce performance in relevant domains. For instance, Mol2vec \citep{Mol2vec}, MAT \citep{pretrained_MAT}, and R-MAT \citep{pretrained_RMAT} were trained on a limited subset of elements, which strongly diminished their performance in ecotoxicology and agrochemistry \citep{ApisTox_ML}. Conversely, pretraining on data that include metals and metalloids is crucial for ML models in areas like synthesis prediction (for example organometallic reactions in Suzuki coupling \citep{suzuki_coupling}) or oncological therapeutics \citep{organometallics_anticancer}. Thus, rich representation of element groups is an important indicator of both quality and diversity of molecular datasets.

Figure \ref{atom_groups} illustrates the composition of element groups in MolPILE compared to other datasets. MolPILE exhibits a broad distribution, with a relatively high fraction of halogen-, metalloid-, and metal-containing compounds. Its constituent datasets vary, but interestingly natural products in SuperNatural3 and COCONUT include far fewer nitrogen- and sulfur-containing compounds. ChEMBL, GDB-17, and ZINC are notably limited in elemental diversity. Specifically, ZINC and GDB-17 contain no metalloids or metals, ChEMBL includes very few, and GDB-17 heavily underrepresents halogens. These patterns suggest that neural networks pretrained on ChEMBL or ZINC may underperform on chemical spaces beyond traditional medicinal chemistry, such as anti-tumor metallodrugs.

\subsection{\#Circles structural diversity}

\begin{figure}
    \centering
    \includegraphics[width=\textwidth]{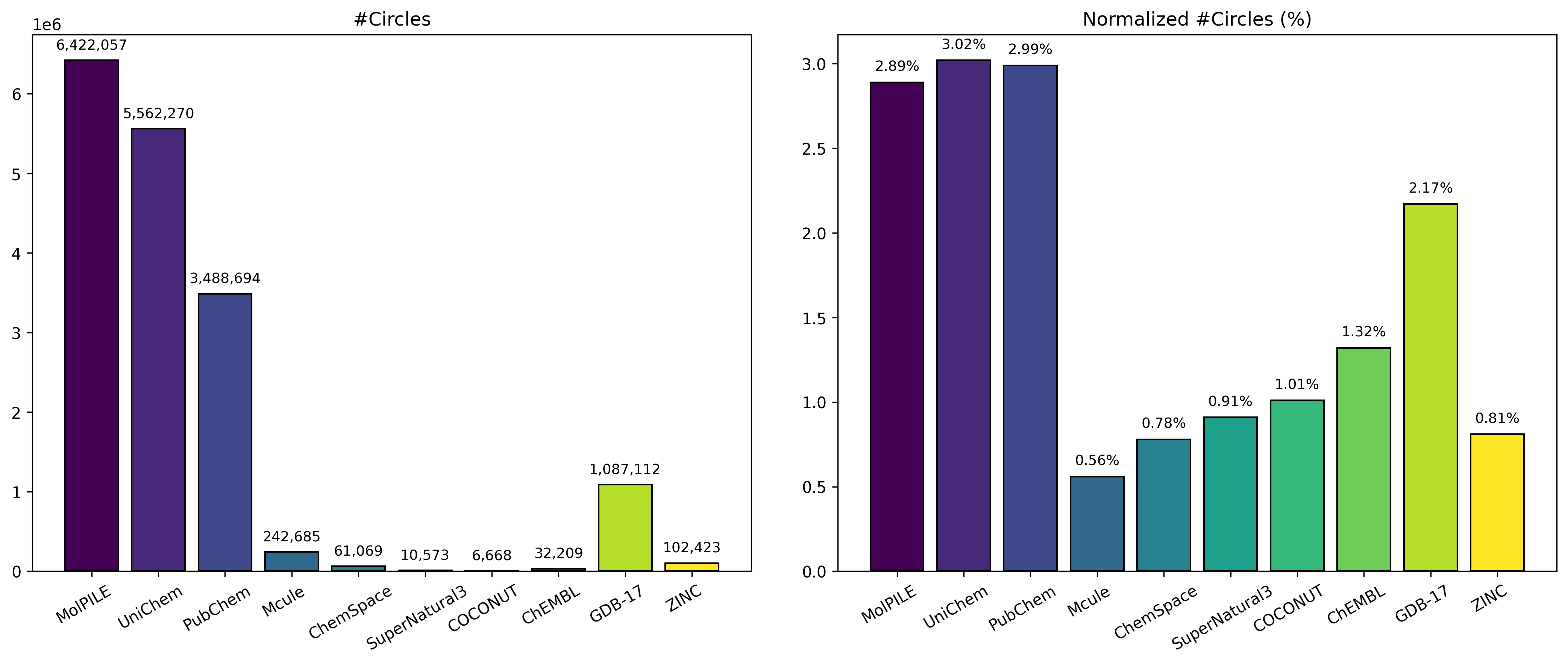}
    \caption{\#Circles of each dataset and its normalized values, accounting for dataset size differences.}
    \label{nCircles}
\end{figure}

Measuring chemical space coverage requires assessing how structurally representative the compounds in a dataset are. While one could enumerate an enormous space of nearly identical compounds by varying scaffolds minimally or substituting peripheral functional groups, such an approach would yield a dataset with intuitively low diversity. To address this, the \#Circles metric was introduced in \cite{nCircles} as a measure of diversity for generated molecular spaces, and it can likewise be applied to general compound datasets \citep{ApisTox}. Formally, it is defined as the maximum number of disjoint circles of radius $t$, with centers located at elements of the molecule set $\mathcal{S}$ (equivalent to the packing number in topology), as shown in Equation \ref{eq_nCircles}.

\begin{equation}\label{eq_nCircles}
\#Circles(\mathcal{S}, d, t) = \max_{\mathcal{C}\subseteq\mathcal{S}} \left| \mathcal{C} \right |\quad \text{where} \quad d(x,y) \ge t \quad  \forall x \neq y \in \mathcal{C} 
\end{equation}

To compare datasets of different sizes and assess their relative diversity, we also compute normalized \#Circles by dividing the raw value by dataset size \citep{ApisTox}. Following the original formulation, we use binary ECFP fingerprints with Tanimoto distance and a threshold of $t = 0.75$. Although exact computation of \#Circles is NP-hard \citep{nCircles_nphard}, it can be efficiently approximated and parallelized \citep{nCircles}.

Figure \ref{nCircles} presents the results for both raw and normalized \#Circles. As expected, the raw metric scales with dataset size, with MolPILE achieving the largest value overall. Normalized \#Circles, however, reveals differences in structural diversity independent of dataset size. Large-scale collections such as MolPILE, UniChem, and PubChem exhibit high diversity, each with values around 3\%. In contrast, widely used ZINC shows remarkably low value of 0.81\%. This is concerning for large-scale model training, as neural scaling laws require sufficiently diverse data to hold. Indeed, this observation may explain the performance saturation reported in ChemFM \citep{ChemFM}, where decoder-only transformers trained on ZINC plateaued starting at just 60M parameters. This phenomenon was not observed with PubChem when using the same number of compounds, likely due to its higher diversity. ChEMBL also shows limited diversity, with a normalized \#Circles of 1.32\%, only slightly better than ZINC.

\subsection{Synthesizability}

The Synthesizability Score (SAScore) \citep{SAScore} summarizes the ease of synthesizing a given compound. It ranges from 1 to 10, with lower values indicating easier synthesis. Compared to alternative scoring systems, SAScore has been shown to yield more accurate and intuitive results \citep{SAScore_comparison,SAScore_comparison_2}. Typically, compounds with SAScore above 5–6 are considered difficult to synthesize using standard techniques. The overall distribution of SAScore in a dataset therefore serves as a proxy for its practical quality, with realistic, experimentally synthesizable datasets expected to exhibit predominantly low values.

Figure \ref{figure_sascore} presents boxplots of SAScore values across datasets (see Appendix \ref{appendix_sascore_table} for full statistics). GDB-17 exhibits markedly higher scores than other datasets, reflecting the general impracticality of synthesizing many enumerated compounds. This raises concerns for pretraining models, which may overfit to molecules that are challenging or unrealistic to access. In contrast, MolPILE shows a reasonable distribution with a median SAScore of 3.05, consistent with its construction from experimentally validated compounds. ChEMBL and ZINC also display sensible distributions. Slightly higher scores and heavy tails in SuperNatural3 and COCONUT are expected, as natural products are generally more difficult to synthesize using traditional methods \citep{SAScore}.

\begin{figure}
    \centering
    \includegraphics[width=0.8\textwidth]{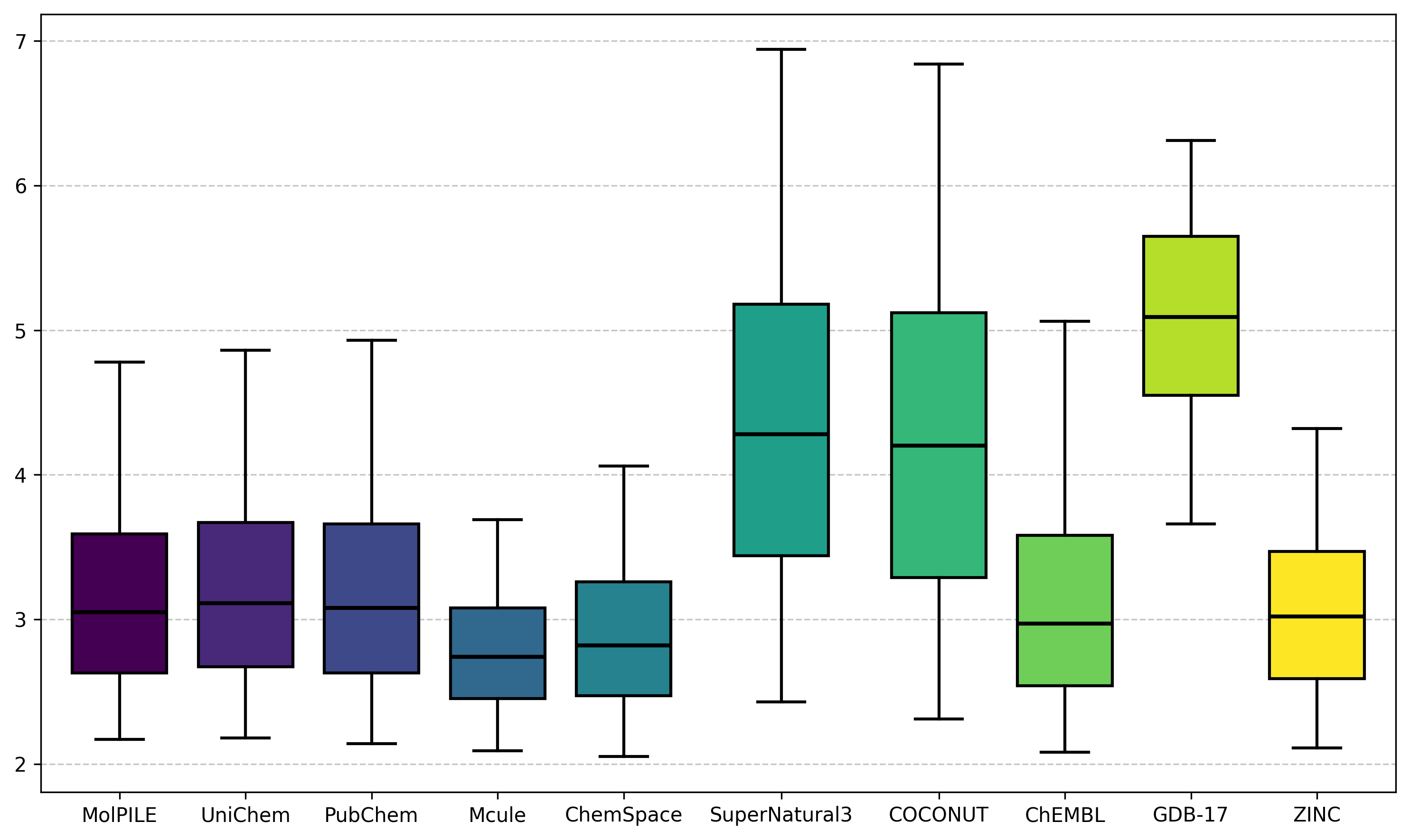}
    \caption{SAScore distribution in each dataset.}
    \label{figure_sascore}
    \vspace{-3mm}
\end{figure}



\subsection{Further analyses}

We performed several additional analyses of MolPILE and alternative datasets. Due to space constraints, we summarize the main findings here and refer readers to the appendices for details.

Appendix \ref{appendix_molecular_descriptors} reports distributions of molecular descriptors used in our feasibility filter (e.g., molecular weight, logP, HBA, HBD). While most datasets follow expected patterns, ZINC stands out as highly homogeneous. It consists almost entirely of typical medicinal compounds consistent with textbook guidelines for orally bioavailable drugs, e.g., low weight, logP about 3, and very few HBD.

In Appendix \ref{appendix_molecular_filters}, we further analyze molecular filters, both physicochemical (e.g., Lipinski's rule of 5) and substructure-based (e.g., REOS). GDB-17 stands out by passing Lipinski's rule in 100\% of cases, as well as GSK and Veber filters. Since molecular filters act as simple rule-based classifiers that restrict chemical space by predefined conditions, this outcome highlights the low diversity of GDB-17, reflecting its conservative enumerated design.

Appendix \ref{appendix_functional_groups_scaffolds_salts} reports diversity analyses in terms of Bemis–Murcko scaffolds \citep{scaffolds}, functional groups detected by Ertl’s algorithm \citep{functional_groups}, and salts. MolPILE exhibits broad chemical diversity, with over 3.6M unique scaffolds, 128k functional groups, and 1M salts. By comparison, GDB-17 and ZINC contain virtually no salts, while ChEMBL and ZINC include fewer than 6k unique functional groups, underscoring the limitations of existing datasets.


\section{Model pretraining}

To practically evaluate the impact of MolPILE qualities, we retrain Mol2vec \citep{Mol2vec} and ChemBERTa \citep{ChemBERTa} with it. We select those models, as they represent two significantly different models and approaches to preparing pretraining data. Further, many of the largest pretrained models for molecules have been SMILES-based, so we focus on this approach. Evaluating large-scale pretraining of graph neural networks on MolPILE is a direction for future work.

Originally, Mol2vec was trained on 20M molecules from ZINC15 and ChEMBL with stringent filtering: molecular weight $\in [12, 600]$, number of heavy atoms $\in [3, 50]$, logP $\in [-5, 7]$, removing counterions and solvents, as well as keeping only elements $[H, B, C, N, O, F, P, S, Cl, Br]$. ChemBERTa was originally trained on 77M molecules from PubChem, chosen randomly without filtering.

We retrained Mol2vec and ChemBERTa (MLM variant) on MolPILE, closely following their original implementations. Inspired by \cite{pretrained_feature_extractors}, we use those models as pretrained feature extractors, training a Random Forest model on top of their embeddings. We include a variety of benchmarks: MoleculeNet \citep{MoleculeNet} and TDC \citep{TDC} for molecular property prediction, WelQrate \citep{WelQrate} for ligand-based virtual screening, ToxBench dataset \citep{ToxBench} for ligand-only binding prediction, as well as ApisTox dataset \citep{ApisTox,ApisTox_ML} representing ecotoxicology. In total, we use 48 individual datasets. We report the metric recommended for each benchmark by its authors: AUROC, MAE, RMSE, or BEDROC \citep{BEDROC}. We report average gain in each benchmark, as well as the number of wins of the retrained model. See Appendix \ref{appendix_model_training_evaluation_datasets} for details on datasets, training procedures, and all results on individual datasets.

Results are summarized in Table \ref{table_retrained_model_gains}, showing that retraining on MolPILE consistently improves performance across the board. Even the relatively simple Mol2vec model, which lacks many parameters to fully exploit an 11 times larger dataset, shows clear gains. This emphasizes the importance of dataset variety and reasonable filtering for generalization. ChemBERTa likewise achieves substantial improvements, particularly on TDC datasets. The only exception is its weaker performance on ApisTox, likely due to the dataset’s atypical and extremely limited vocabulary of just 600 tokens. To benefit from larger and more diverse datasets, ChemBERTa would need to scale up the tokenizer. In contrast, Mol2vec, which does not constrain vocabulary size, achieves strong gains on this agrochemical dataset. Notably, both models perform much better on MoleculeNet, TDC, and WelQrate, indicating that greater data quantity and diversity also benefit ADMET and virtual screening tasks.

\begin{table}[h]
\caption{Results of original and retrained Mol2vec and ChemBERTa.}
\resizebox{\textwidth}{!}{
\begin{tabular}{ccccccccc}
\hline
\multirow{2}{*}{\textbf{Model}} & \multirow{2}{*}{\textbf{Measure}} & \multicolumn{2}{c}{\textbf{MoleculeNet}} & \multicolumn{2}{c}{\textbf{TDC}} & \textbf{WelQrate} & \textbf{ToxBench} & \textbf{ApisTox} \\
 &  & \textbf{AUROC $\uparrow$} & \textbf{MAE $\downarrow$} & \textbf{AUROC $\uparrow$} & \textbf{MAE $\downarrow$} & \textbf{BEDROC $\uparrow$} & \textbf{RMSE $\downarrow$} & \textbf{AUROC $\uparrow$} \\ \hline
\multirow{2}{*}{Mol2vec} & Avg. gain & +1.79 & -0.203 & +0.58 & -0.496 & +1.52 & -0.139 & +2.23 \\ 
 & \# of wins & 3/8 & 3/3 & 15/18 & 4/8 & 7/9 & 1/1 & 1/1 \\
 \hline
\multirow{2}{*}{ChemBERTa} & Avg. gain & +1.09 & -0.200 & +1.26 & -1.417 & +1.07 & -0.017 & -2.81 \\ 
 & \# of wins & 6/8 & 3/3 & 13/18 & 5/8 & 5/9 & 1/1 & 0/1 \\
 \hline
\end{tabular}
}
\label{table_retrained_model_gains}
\end{table}

\section{Conclusions}

In this work, we introduced MolPILE, a large-scale dataset of nearly 222 million small molecules, containing experimentally validated and synthesizable compounds. Through a multi-step workflow, we ensured high-quality, deduplicated, and standardized molecular structures. Our analyses demonstrate that MolPILE offers high diversity, broad chemical space coverage, and superior structural quality compared to widely used datasets such as ChEMBL, GDB-17, and ZINC. We further show that these qualities translate into improved generalization of ML models, with Mol2vec and ChemBERTa pretrained on MolPILE achieving superior performance. MolPILE is released in SMILES format, along with curated, diversity-focused subsets, to facilitate accessible benchmarking. We believe MolPILE will support the development of higher-quality and broadly applicable molecular representation learning models, and serve as a standardized resource similar to ImageNet and PILE.

\section{Reproducibility statement}

All code for reproducing the results is provided in the supplementary material. Scripts for creating MolPILE work end-to-end, downloading and processing the data. Similarly, we include scripts for working with ChEMBL, GDB-17, and ZINC. Code also contains scripts to retrain Mol2vec and ChemBERTa, and evaluate them on benchmarks. All dependencies are managed with \texttt{uv} manager, and we also include \texttt{uv.lock} file with all direct and transitive dependency versions. Instructions for setup and running all elements are included in the README. All analyses presented as plots or summaries in the main body have been expanded and detailed in appendices, including full results tables. Additional descriptions of datasets, models, algorithms (including diverse subset picking for MolPILE subsets), and implementation and hardware used, are included in appendices.

\section*{Acknowledgements}

We thank BIT Student Scientific Group for computational resources and aid for this project. This work was supported by the Ministry of Science and Higher Education funds assigned to AGH University in Krakow. We thank Aleksandra Elbakyan for her work and support for accessibility of science.



\bibliography{bibliography}
\bibliographystyle{iclr2026_conference}
\clearpage

\appendix
\section{Pretraining datasets of current models}
\label{appendix_model_pretraining_datasets}

\begin{table}[h]
\centering
\caption{Overview of pretraining datasets of molecular representation learning models.}
\resizebox{\textwidth}{!}{
\begin{tabular}{|c|c|c|c|c|}
\hline
\textbf{\begin{tabular}[c]{@{}c@{}}Model\\ group\end{tabular}} & \textbf{Model} & \textbf{Data source(s)} & \textbf{Dataset size} & \textbf{Data filtering} \\ \hline
\multirow{4}{*}{\begin{tabular}[c]{@{}c@{}}Graph\\ neural\\ networks\\ (GNNs)\end{tabular}} 
 & \begin{tabular}[c]{@{}c@{}}ContextPred \\ \citep{pretraining_GNNs_GIN}\end{tabular} & \begin{tabular}[c]{@{}c@{}}ZINC / ChEMBL\\ (two stages)\end{tabular} & \begin{tabular}[c]{@{}c@{}}2M / 456k\\ (two stages)\end{tabular} & - \\ \cline{2-5} 
 & \begin{tabular}[c]{@{}c@{}}GEM \\ \citep{GEM}\end{tabular} & ZINC & 20M & - \\ \cline{2-5} 
 & \begin{tabular}[c]{@{}c@{}}GraphMVP \\ \citep{GraphMVP}\end{tabular} & GEOM & 50k & - \\ \cline{2-5} 
 & \begin{tabular}[c]{@{}c@{}}MolR \\ \citep{MolR}\end{tabular} & USPTO & 479k & - \\ \hline
\multirow{4}{*}{\begin{tabular}[c]{@{}c@{}}Graph\\ transformers\end{tabular}} 
 & \begin{tabular}[c]{@{}c@{}}GROVER \\ \citep{GROVER}\end{tabular} & ZINC + ChEMBL & 11M & - \\ \cline{2-5} 
 & \begin{tabular}[c]{@{}c@{}}MAT \\ \citep{pretrained_MAT}\end{tabular} & ZINC & 2M & - \\ \cline{2-5} 
 & \begin{tabular}[c]{@{}c@{}}R-MAT \\ \citep{pretrained_RMAT}\end{tabular} & ZINC + ChEMBL & 4M & \begin{tabular}[c]{@{}c@{}}Lipinski's rule of 5\\ (reduced data from 10M to 4M)\end{tabular} \\ \cline{2-5} 
 & \begin{tabular}[c]{@{}c@{}}Uni-Mol \\ \citep{UniMol}\end{tabular} & ZINC + ChEMBL & 19M & - \\ \hline
\multirow{7}{*}{\begin{tabular}[c]{@{}c@{}}SMILES\\ transformers\end{tabular}} 
 & \begin{tabular}[c]{@{}c@{}}CDDD \\ \citep{CDDD}\end{tabular} & ZINC + PubChem & 72M & \begin{tabular}[c]{@{}c@{}}organic molecules only\\ molecular weight $\in [12, 600]$\\ heavy atoms $\geq 3$\\ logP $\in [-7, 5]$\\ removed steochemistry\\ salts stripped to largest fragments\end{tabular} \\ \cline{2-5} 
 & \begin{tabular}[c]{@{}c@{}}ChemBERTa \\ \citep{ChemBERTa}\end{tabular} & PubChem & \begin{tabular}[c]{@{}c@{}}5M / 10M / 77M\\ (three variants)\end{tabular} & - \\ \cline{2-5} 
 & \begin{tabular}[c]{@{}c@{}}ChemFM \\ \citep{ChemFM}\end{tabular} & UniChem & 178M & - \\ \cline{2-5} 
 & \begin{tabular}[c]{@{}c@{}}Chemformer \\ \citep{Chemformer}\end{tabular} & ZINC & 100M & \begin{tabular}[c]{@{}c@{}}reactivity "reactive"\\ purchasability "annotated"\\ molecular weight $\leq$ 500\\ logP $\leq 5$\end{tabular} \\ \cline{2-5} 
 & \begin{tabular}[c]{@{}c@{}}MolBERT \\ \citep{MolBERT}\end{tabular} & \begin{tabular}[c]{@{}c@{}}GuacaMol\\ (ChEMBL subset)\end{tabular} & 1.6M & - \\ \cline{2-5} 
 & \begin{tabular}[c]{@{}c@{}}MoLFormer \\ \citep{MolFormer}\end{tabular} & ZINC + PubChem & $\sim$1.11B & SMILES length $\leq 211$ \\ \cline{2-5} 
 & \begin{tabular}[c]{@{}c@{}}SimSon \\ \citep{SimSon}\end{tabular} & PubChem & 1M & - \\ \hline
\multirow{2}{*}{\begin{tabular}[c]{@{}c@{}}SELFIES\\ transformers\end{tabular}} 
 & \begin{tabular}[c]{@{}c@{}}ChemGPT \\ \citep{ChemGPT}\end{tabular} & PubChem & 10M & - \\ \cline{2-5} 
 & \begin{tabular}[c]{@{}c@{}}SELFormer \\ \citep{SELFormer}\end{tabular} & ChEMBL & 2M & - \\ \hline
\multirow{3}{*}{\begin{tabular}[c]{@{}c@{}}Hybrid\\ models\end{tabular}} 
 & \begin{tabular}[c]{@{}c@{}}CLAMP \\ \citep{CLAMP}\end{tabular} & PubChem Assays & 2.1M & - \\ \cline{2-5} 
 & \begin{tabular}[c]{@{}c@{}}COATI \\ \citep{COATI}\end{tabular} & \begin{tabular}[c]{@{}c@{}}ChEMBL\\ + GEOM-Drugs\\ + TensorMol\\ + Mcule\\ + ZINC22\\ + Enamine\\ building blocks\end{tabular} & 159M & - \\ \cline{2-5} 
 & \begin{tabular}[c]{@{}c@{}}Mol2vec \\ \citep{Mol2vec}\end{tabular} & ZINC + ChEMBL & 20M & \begin{tabular}[c]{@{}c@{}}RDKit parseable\\ molecular weight $\in [12, 600]$\\ heavy atoms count $\in [3, 50]$\\ logP $\in [-5, 7]$\\ counterions removed\\ solvents removed\\ only elements $[H, B, C, N, O$, \\ $F, P, S, Cl, Br]$\end{tabular} \\ \hline
\end{tabular}
}
\label{appendix_table_models_pretraining}
\end{table}

Table \ref{appendix_table_models_pretraining} summarizes 21 molecular representation learning models and their pretraining datasets, including data sources, sizes (number of molecules), and filtering rules (if any), as selected from the benchmarking study of \cite{pretrained_feature_extractors}.

A large majority of models - 16 out of 21 - apply no filtering, thereby including many low-quality, atypical, duplicated, or erroneous molecules, which MolPILE explicitly removes. Our analyses show that these issues are particularly prevalent in PubChem, UniChem, and ZINC, at least one of which is used by 15 of the 21 models. By contrast, models such as Mol2vec, R-MAT, CDDD, and Chemformer apply strict filtering, but in doing so heavily restrict the explored chemical space. Notably, three models trained on the largest and most diverse datasets - MoLFormer, ChemFM, and COATI - apply no filtering at all, leaving structure viability and synthesizability unchecked. Furthermore, 9 models rely on datasets of fewer than 10 million molecules, which is insufficient to adequately represent chemical space and limits generalization capability.

Taken together, these observations reveal three major shortcomings in existing pretraining datasets: small size, low quality, and low diversity. Many models are trained on datasets too small to generalize broadly across chemistry; others rely on unfiltered datasets containing unrealistic, low-quality molecules; and a third group applies overly strict filtering, eliminating essential diversity. MolPILE addresses all three issues by providing a large-scale, high-quality, and diverse dataset, thereby meeting the key desiderata for pretraining in molecular representation learning.

\section{UniChem and PubChem sources}
\label{appendix_unichem_pubchem_sources}

Here, we provide an overview of the major databases contained within UniChem and PubChem. As listing all sources would be impractical, since both integrate data from hundreds of vendors, we highlight only the most significant ones. Because UniChem and PubChem are incorporated into MolPILE, their sources are included implicitly and do not need to be added separately to our pipeline. Although some sources appear in both databases, MolPILE's standardization and merging steps ensure proper handling of duplicates. Importantly, many UniChem sources are subject to strong filtering, and their inclusion within the MolPILE pipeline introduces numerous additional compounds.

Ten largest UniChem sources:
\begin{enumerate}
    \item PubChem (subset)
    \item Mcule (subset)
    \item SureChEMBL
    \item ZINC (subset)
    \item eMolecules
    \item Nikkaji
    \item ChEMBL
    \item BindingDB
    \item EPA CompTox Dashboard
    \item SwissLipids
\end{enumerate}

Ten largest PubChem sources:
\begin{enumerate}
    \item Aurora Fine Chemicals LLC
    \item SureChEMBL
    \item AKos Consulting \& Solutions
    \item PATENTSCOPE (WIPO)
    \item Google Patents
    \item IBM
    \item ChemSpider (subset)
    \item ZINC (subset)
    \item DiscoveryGate
    \item NextBio
\end{enumerate}

Other sources of PubChem also include subsets from, e.g., Mcule, Enamine, ChemSpace, MolGenie, ChemDB, ChEMBL, ChemBank, Wikidata, eMolecules.

\section{MolPILE pipeline implementation and hardware}
\label{appendix_implementation_and_hardware}

For implementing MolPILE, we use:
\begin{itemize}
    \item uv for dependency management
    \item aria2 and rapidgzip \citep{rapidgzip} for files downloading and unpacking
    \item Joblib for multiprocessing
    \item Polars \citep{Polars} and DuckDB \citep{DuckDB} for data engineering and files interaction
    \item RDKit \citep{RDKit} and scikit-fingerprints \citep{scikit-fingerprints} for chemoinformatics tasks
    \item Gensim \citep{Gensim} for implementing Mol2vec training
    \item HuggingFace Transformers \citep{HuggingFace_transformers} and Datasets \citep{HuggingFace_datasets} for implementing ChemBERTa training
    \item scikit-learn \citep{scikit-learn} for implementing and evaluating models and benchmarks
\end{itemize}

For calculating MolPILE and performing analyses, we used a server with 128-core CPU and 512 GB RAM. We have also validated that it works on an average-grade machine with 12-core CPU, 64 GB RAM, and additional swap memory. The entire end-to-end pipeline, including downloading source datasets, takes under 12 hours on a server, and under 48 hours on the average-grade PC.

\section{Feasibility filter design}
\label{appendix_feasibility_filter_design}

Design of \textit{molecular feasibility filter} included three main components: analysis of distributions of descriptors in source datasets, consulting an expert chemist, and referencing the literature. We also compared those distributions to values of approved drugs from DrugBank. To recall from the main body, the final molecular feasibility filter criteria are:
\begin{itemize}
    \item molecular fragments $\leq 3$
    \item length of InChI $< 2000$
    \item molecular weight $\leq 2500$
    \item number of atoms $\leq 150$
    \item HBA $\leq 20$
    \item HBD $\leq 15$
    \item logP in range $[-10, 25]$
    \item TPSA $\leq 500$
    \item number of rotatable bonds $\leq 60$
\end{itemize}

We primarily focused on analyzing the basic distribution statistics like min/max, quartiles, or very low/high percentiles, as this information summarizes the knowledge about tails of the distribution. Similar techniques are often used to design molecular filters \citep{}, but they are typically more specific, leaving only, e.g., lead-like molecules \citep{} or pesticides of specific type \citep{}. Here, we are interested in filtering out most extreme and unusual molecules based on that data, summarized in:
\begin{itemize}
    \item molecular weight - Table \ref{appendix_table_molwt}
    \item number of atoms - Table \ref{appendix_table_num_atoms}
    \item hydrogen bond acceptors (HBA) - Table \ref{appendix_table_hba}
    \item hydrogen bond donors (HBD) - Table \ref{appendix_table_hbd}
    \item water-octanol partition coefficient logarithm (logP) - Table \ref{appendix_table_logp}
    \item topological polar surface area (TPSA) - Table \ref{appendix_table_tpsa}
    \item number of rotatable bonds - Table \ref{appendix_table_num_rot_bonds}
\end{itemize}

For readability, numbers are presented as integers where possible without loss of precision (HBA, HBD, number of rotatable bonds).

First, we limit the number of molecular fragments to at most $3$ to keep only typical single-component molecules and salts. Complexes with over 3 fragments are extremely rare, atypical, and very unstable. In most cases, chemists wouldn't consider them ``molecules'' for the purpose of data processing, but rather complexes or mixtures.

We impose a few size limits to remove proteins, peptides and other polymers, large crystals, and metal complexes. This removes compounds too large to be considered ``small molecules'', which are of primary interest here. While peptide-based therapeutics are of interest, they require dedicated processing, algorithms, and datasets, incorporating more informatics specific for biologics and proteins \citep{filtering_peptides}. Those conditions are length of InChI $< 2000$, molecular weight $\leq 2500$, and number of atoms $\leq 150$, consistent with literature \citep{FAF4,filtering_databases}. Note that those values are larger than 99th percentile of all source datasets, thus removing compounds only from the 1\% of the most extreme outliers. Analyzing the DrugBank approved drugs, the 99th percentile of molecular weight and number of atoms are $1565.01$ and $103$, respectively, so we also cover those \citep{filtering_databases}. We also note that chemical LLMs based on SMILES or SELFIES typically impose the limit for sequence length of 512 tokens, due to quadratic complexity of attention mechanism \citep{ChemBERTa,Chemformer,SELFormer}. Thus, removing those larger compounds ensures proper data quality and removing outliers, while having minimal impact on its size.

Another group of requirements is related to compound stability and synthesizability. We do not want to include molecules that are clearly too unstable and unobtainable under typical synthesis routes, which also negatively impacts their bioavailability, selectivity, cell permeability, and protein binding properties \citep{filters_Veber,HBD_impact,HBA_HBD_impact}. In extreme cases, molecules would not have any stable structure at all under viable conditions, making their behavior unpredictable and compound unusable in practice. Thus, we limit the hydrogen bond acceptors (HBA) to at most $20$, HBD at most $15$, and number of rotatable bonds $\leq 60$. Those requirements still allow very flexible molecules, covering 95th percentile of DrugBank allowed drugs (HBA $14$, HBD $7$, rotatable bonds $16$), but we also allow larger values due to quite flexible RDKit definition of rotatable bonds.

We limit the TPSA to at most 500 to allow almost all natural products from SuperNatural3 and COCONUT to be included. This number also admits large industrial chemistry compounds and various molecules outside medicinal chemistry, not aiming to permeate cell barriers. The main pharmaceutical motivation for TPSA-based filtering is to ensure bioavailability and cell permeability, with even stricter requirements for more specialized drugs like those targeting central nervous system, which are recommended to be under 90 TPSA to permeate the blood-brain barrier \citep{TPSA_1, TPSA_2}. As MolPILE aims to be more general, covering not only medicinal chemistry, we allow quite large TPSA.

The logarithm of the water-octanol partition coefficient, logP, measures the compound lipophilicity, and is one of the key indicators of its potential ability to cross biological barriers. The typical range of values conducive to biological activity is between approximately 0 and 5, with high values indicating low solubility in an aqueous environment, and conversely low values suggesting limited ability to penetrate lipid membranes. It can usually be well-approximated computationally by a simple atomic contributions model of \citep{logP_Wildman_Crippen}, and it is the only method available in RDKit. This method generally works well and for a wide variety of compounds. However, one should note that it has been designed primarily for small organic molecules, and its results may be subject to greater error outside this applicability domain, e.g. for larger biomolecules or specific classes of chemical compounds, such as polymers or inorganic compounds. logP values of 10 or higher are extremely rare, and compounds with such characteristics typically have no practical application as biologically active substances. Extreme values, such as logP values around 20, should generally be treated as computational artifacts rather than actual physicochemical parameters. Similar observations can be made for negative values, and those limitations also stem from physical limitations of experimental methods \citep{logp_measurement}. For those reasons, taking into account both the distribution of values of this descriptor and the reasonable limits observed in typical chemical measurements, a range of analysis from -10 to 25 was adopted.

\begin{table}[H]
\centering
\caption{Molecular weight distributions statistics.}
\resizebox{\textwidth}{!}{
\begin{tabular}{|l|r|r|r|r|r|r|r|r|r|r|}
\hline
\textbf{Dataset} & \textbf{min} & \textbf{p1} & \textbf{p5} & \textbf{Q1} & \textbf{mean} & \textbf{median} & \textbf{Q3} & \textbf{p95} & \textbf{p99} & \textbf{max} \\ \hline
UniChem & 1.0 & 154.0 & 205.3 & 289.4 & 407.9 & 356.4 & 435.0 & 725.9 & 1612.8 & 113821.4 \\ \hline
PubChem & 1.0 & 151.2 & 200.4 & 280.4 & 429.8 & 358.4 & 461.5 & 821.7 & 2092.0 & 113821.4 \\ \hline
Mcule & 9.0 & 188.6 & 250.3 & 345.4 & 395.5 & 393.5 & 448.5 & 528.6 & 595.7 & 6179.4 \\ \hline
ChemSpace & 13.0 & 189.2 & 247.3 & 314.2 & 360.9 & 352.4 & 401.5 & 488.4 & 580.5 & 8073.5 \\ \hline
SuperNatural3 & 1.0 & 150.2 & 222.4 & 345.4 & 512.0 & 439.6 & 600.8 & 1013.6 & 1444.8 & 5033.8 \\ \hline
COCONUT & 1.0 & 136.2 & 210.3 & 334.4 & 506.6 & 431.5 & 601.6 & 1009.1 & 1435.8 & 7860.7 \\ \hline
\end{tabular}
}
\label{appendix_table_molwt}
\end{table}

\begin{table}[H]
\centering
\caption{Number of atoms distributions statistics.}
\begin{tabular}{|l|r|r|r|r|r|r|r|r|r|r|}
\hline
\textbf{Dataset} & \textbf{min} & \textbf{p1} & \textbf{p5} & \textbf{Q1} & \textbf{mean} & \textbf{median} & \textbf{Q3} & \textbf{p95} & \textbf{p99} & \textbf{max} \\ \hline
UniChem & 1 & 10 & 14 & 20 & 28.2 & 25 & 30 & 51 & 113 & 999 \\ \hline
PubChem & 1 & 10 & 13 & 19 & 29.7 & 25 & 32 & 59 & 148 & 910 \\ \hline
Mcule & 1 & 13 & 17 & 24 & 28.1 & 28 & 32 & 38 & 43 & 419 \\ \hline
ChemSpace & 1 & 13 & 17 & 22 & 25.2 & 25 & 28 & 34 & 40 & 566 \\ \hline
SuperNatural3 & 1 & 10 & 16 & 25 & 36.4 & 31 & 43 & 72 & 100 & 352 \\ \hline
COCONUT & 1 & 10 & 15 & 24 & 36.0 & 31 & 43 & 71 & 100 & 551 \\ \hline
\end{tabular}
\label{appendix_table_num_atoms}
\end{table}

\begin{table}[H]
\centering
\caption{Hydrogen bond acceptors (HBA) distributions statistics.}
\begin{tabular}{|l|r|r|r|r|r|r|r|r|r|r|}
\hline
\textbf{Dataset} & \textbf{min} & \textbf{p1} & \textbf{p5} & \textbf{Q1} & \textbf{mean} & \textbf{median} & \textbf{Q3} & \textbf{p95} & \textbf{p99} & \textbf{max} \\ \hline
UniChem & 0 & 0 & 2 & 3 & 4.8 & 4 & 6 & 9 & 19 & 608 \\ \hline
PubChem & 0 & 0 & 1 & 3 & 5.0 & 4 & 6 & 10 & 23 & 729 \\ \hline
Mcule & 0 & 1 & 2 & 3 & 4.7 & 5 & 6 & 8 & 9 & 191 \\ \hline
ChemSpace & 0 & 1 & 2 & 4 & 4.8 & 5 & 6 & 8 & 9 & 121 \\ \hline
SuperNatural3 & 0 & 1 & 2 & 4 & 7.1 & 6 & 8 & 17 & 28 & 106 \\ \hline
COCONUT & 0 & 1 & 2 & 4 & 7.0 & 6 & 8 & 17 & 28 & 191 \\ \hline
\end{tabular}
\label{appendix_table_hba}
\end{table}

\begin{table}[H]
\centering
\caption{Hydrogen bond donors (HBD) distributions statistics.}
\begin{tabular}{|l|r|r|r|r|r|r|r|r|r|r|}
\hline
\textbf{Dataset} & \textbf{min} & \textbf{p1} & \textbf{p5} & \textbf{Q1} & \textbf{mean} & \textbf{median} & \textbf{Q3} & \textbf{p95} & \textbf{p99} & \textbf{max} \\ \hline
UniChem & 0 & 0 & 0 & 1 & 1.5 & 1 & 2 & 4 & 8 & 444 \\ \hline
PubChem & 0 & 0 & 0 & 1 & 1.6 & 1 & 2 & 4 & 9 & 292 \\ \hline
Mcule & 0 & 0 & 0 & 1 & 1.1 & 1 & 1 & 3 & 4 & 116 \\ \hline
ChemSpace & 0 & 0 & 0 & 1 & 1.2 & 1 & 2 & 3 & 4 & 122 \\ \hline
SuperNatural3 & 0 & 0 & 0 & 1 & 3.0 & 2 & 4 & 10 & 17 & 80 \\ \hline
COCONUT & 0 & 0 & 0 & 1 & 2.9 & 2 & 4 & 10 & 17 & 119 \\ \hline
\end{tabular}
\label{appendix_table_hbd}
\end{table}

\begin{table}[H]
\centering
\caption{Water-octanol partition coefficient logarithm (logP) distributions statistics.}
\begin{tabular}{|l|r|r|r|r|r|r|r|r|r|r|}
\hline
\textbf{Dataset} & \textbf{min} & \textbf{p1} & \textbf{p5} & \textbf{Q1} & \textbf{mean} & \textbf{median} & \textbf{Q3} & \textbf{p95} & \textbf{p99} & \textbf{max} \\ \hline
UniChem & -2048.5 & -1.3 & 0.5 & 2.2 & 3.9 & 3.3 & 4.6 & 8.4 & 20.7 & 508.3 \\ \hline
PubChem & -2234.1 & -1.5 & 0.5 & 2.2 & 4.3 & 3.5 & 4.9 & 10.3 & 24.7 & 709.2 \\ \hline
Mcule & -83.7 & 0.3 & 1.5 & 3.1 & 4.1 & 4.1 & 5.1 & 6.4 & 7.5 & 61.4 \\ \hline
ChemSpace & -49.1 & -0.1 & 0.9 & 2.2 & 3.2 & 3.2 & 4.1 & 5.6 & 6.9 & 47.3 \\ \hline
SuperNatural3 & -33.3 & -4.2 & -1.3 & 1.9 & 4.2 & 3.6 & 5.3 & 13.5 & 20.2 & 47.6 \\ \hline
COCONUT & -83.7 & -4.5 & -1.4 & 1.8 & 4.3 & 3.5 & 5.3 & 16.2 & 20.7 & 77.1 \\ \hline
\end{tabular}
\label{appendix_table_logp}
\end{table}

\begin{table}[H]
\centering
\caption{Topological polar surface area (TPSA) distributions statistics.}
\resizebox{\textwidth}{!}{
\begin{tabular}{|l|r|r|r|r|r|r|r|r|r|r|}
\hline
\textbf{Dataset} & \textbf{min} & \textbf{p1} & \textbf{p5} & \textbf{Q1} & \textbf{mean} & \textbf{median} & \textbf{Q3} & \textbf{p95} & \textbf{p99} & \textbf{max} \\ \hline
UniChem & 0.0 & 0.0 & 20.2 & 49.3 & 78.4 & 69.4 & 91.7 & 151.9 & 323.7 & 16013.0 \\ \hline
PubChem & 0.0 & 0.0 & 16.4 & 45.7 & 80.1 & 68.1 & 93.5 & 165.8 & 388.8 & 11400.0 \\ \hline
Mcule & 0.0 & 18.8 & 33.1 & 55.6 & 73.7 & 71.5 & 89.4 & 118.8 & 145.6 & 3038.9 \\ \hline
ChemSpace & 0.0 & 21.1 & 35.0 & 56.5 & 74.4 & 72.4 & 89.4 & 118.5 & 146.1 & 3537.2 \\ \hline
SuperNatural3 & 0.0 & 9.2 & 26.3 & 63.2 & 117.7 & 89.9 & 140.6 & 296.1 & 500.1 & 2231.2 \\ \hline
COCONUT & 0.0 & 4.4 & 26.3 & 61.8 & 116.1 & 86.2 & 136.7 & 297.6 & 503.7 & 3548.4 \\ \hline
\end{tabular}
}
\label{appendix_table_tpsa}
\end{table}

\begin{table}[H]
\centering
\caption{Number of rotatable bonds distributions statistics.}
\begin{tabular}{|l|r|r|r|r|r|r|r|r|r|r|}
\hline
\textbf{Dataset} & \textbf{min} & \textbf{p1} & \textbf{p5} & \textbf{Q1} & \textbf{mean} & \textbf{median} & \textbf{Q3} & \textbf{p95} & \textbf{p99} & \textbf{max} \\ \hline
UniChem & 0 & 0 & 1 & 3 & 6.4 & 5 & 7 & 13 & 43 & 784 \\ \hline
PubChem & 0 & 0 & 1 & 3 & 6.6 & 5 & 7 & 15 & 46 & 648 \\ \hline
Mcule & 0 & 1 & 2 & 4 & 5.6 & 5 & 7 & 10 & 13 & 228 \\ \hline
ChemSpace & 0 & 1 & 2 & 3 & 4.8 & 5 & 6 & 8 & 10 & 200 \\ \hline
SuperNatural3 & 0 & 0 & 1 & 3 & 9.0 & 5 & 9 & 41 & 55 & 148 \\ \hline
COCONUT & 0 & 0 & 0 & 3 & 9.4 & 5 & 9 & 46 & 57 & 224 \\ \hline
\end{tabular}
\label{appendix_table_num_rot_bonds}
\end{table}

\section{MolPILE source datasets contributions}
\label{appendix_source_datasets_contributions}

\begin{table}[h]
\caption{Cross-dataset counts after filtering.}
\setlength{\tabcolsep}{6pt}
\resizebox{\textwidth}{!}{
\begin{tabular}{lrrrrrr}
\toprule
\textbf{Dataset} & \textbf{UniChem} & \textbf{PubChem} & \textbf{Mcule} & \textbf{ChemSpace} & \textbf{SuperNatural3} & \textbf{COCONUT} \\
\midrule
UniChem       & 0         & 72636030  & 172822096 & 176788176 & 182865354 & 183310907 \\
PubChem       & 6068050   & 0         & 106906633 & 110323432 & 116332956 & 116765767 \\
Mcule         & 32361592  & 33014109  & 0         & 40134274  & 43262835  & 43324997 \\
ChemSpace     & 726029    & 829265    & 4532631   & 0         & 7737809   & 7773204 \\
SuperNatural3 & 100482    & 136064    & 958467    & 1035084   & 0         & 590154 \\
COCONUT       & 48647     & 71487     & 523241    & 573091    & 92766     & 0 \\
\bottomrule
\end{tabular}
}
\label{appendix_table_cross_dataset_gains}
\end{table}

In Table \ref{appendix_table_cross_dataset_gains}, we present a cross-dataset new molecule counts. In $i$-th row and $j$-th column, we show the number of molecules present in dataset $i$, but not in $j$-th. Each pair of datasets contains a non-overlapping subset, meaning that each database brings further new molecules to MolPILE.

\section{Licensing}
\label{appendix_licensing}

MolPILE is a collection of processed datasets, that we redistribute. It is shared as a single Parquet file with columns \textit{source} and \textit{ID}, where source is e.g. ``PubChem'' or ``UniChem'', and ID is the original identifier in the given database, e.g. PubChem CID. Each source has its own separate license, which we list below. As its entirety, MolPILE does not have a single license, as it is a collection, not a single dataset. Users interested in that can easily filter the dataset by source. In case of PubChem and UniChem, users may also want to check the individual licenses of their sources. Users using those sources are also asked to cite the appropriate publications. We do not make any claims about licensing of models trained on MolPILE, nor put any additional limitations.

Licensing of source datasets:
\begin{itemize}
    \item PubChem \citep{PubChem} - CC0 (public domain)
    \item UniChem \citep{UniChem} - CC-BY-4.0
    \item Mcule \citep{Mcule} - CC-BY-NC-4.0
    \item ChemSpace \citep{ChemSpace} - CC-BY-NC-4.0
    \item SuperNatural3 \citep{SuperNatural3} - not specified, only ``freely available''
    \item COCONUT \citep{COCONUT} - CC0 (public domain)
\end{itemize}

\section{Diverse subsets picking}
\label{appendix_diversity_subsets}

The complete MolPILE dataset contains 222 million molecules, which may be too large for initial experimentation or for training computationally expensive models, such as those based on contrastive learning. Selecting a random subset would sample proportionally to the local density of chemical space, thereby potentially biasing the model toward more easily synthesizable compounds, primarily from synthetic medicinal chemistry. To address this, and inspired by diversity selection methods used in virtual screening, we designed a procedure to construct diverse subsets. Specifically, we generated subsets of 100K, 1M, 5M, and 10M molecules.

The algorithm is based on prototype clustering using maximum diversity picking. Formally, the goal is to select a subset of $M$ compounds from a dataset of size $N$. First, $K$ compounds are chosen as cluster centers using the maximum diversity picking \citep{maximum_diversity_picking} algorithm with minimum distance $t$. This ensures that the subset achieves maximal diversity, distributing the centers uniformly across the chemical space, so that the sum of their pairwise distances is maximized. Next, each of the remaining $N-K$ molecules is assigned to the nearest cluster center based on the Tanimoto distance. From each cluster, we select its center and randomly choose $\lfloor \frac{M-K}{K} \rfloor - 1$ additional molecules. If each cluster contains at least $\frac{M-K}{K}$ compounds, the algorithm terminates. In cases of highly uneven data density, where some clusters are much smaller than others, the subset may not reach size $M$. In such cases, the remaining molecules are randomly sampled from the unselected pool until the subset reaches the desired size.

As maximum diversity picking problem is NP-hard \citep{maximum_diversity_picking_NP_hard}, we use the MaxMin approximation available in RDKit \citep{MaxMin_RDKit}. A distance threshold of $t=0.9$ is applied, ensuring a high degree of separation of cluster centers. Under these settings, MolPILE contains $K=7457$ cluster centers.

To verify the impact of this procedure, we compare it to randomly selected subsets of the same size. To quantify diversity, we compute the distribution of pairwise Tanimoto distances among molecules within each subset. For computational efficiency, we sample 100 million pairs to approximate the distribution, which should be sufficient; for example, authors of \cite{Tanimoto_distribution} demonstrated that typically just 1 million pairs is sufficient to approximate well. We summarize the distributions using percentiles, mean, and median, and additionally compute the Wasserstein distance and the Kolmogorov-Smirnov (K-S) statistic between the two distributions. Note that larger values indicate that molecules are more distant. Subset statistics are reported in Table \ref{table_appendix_diverse_subset_100k} for 100K, Table \ref{table_appendix_diverse_subset_1M} for 1M, Table \ref{table_appendix_diverse_subset_5M} for 5M, and Table \ref{table_appendix_diverse_subset_10M} for 10M molecules.

The diverse subsets exhibit higher inter-molecule Tanimoto distances across the entire distribution. This is further supported by the relatively high Wasserstein distance (bounded by 1, like the Tanimoto distance) and the very high K-S statistic. Together, these results indicate that the diverse subsets closely approximate the highly diverse nature of MolPILE and should therefore be preferred over random subsets. Using standardized subsets further increases reproducibility of results for models trained on MolPILE.

\begin{table}[]
\centering
\setlength{\belowcaptionskip}{2pt}
\caption{Tanimoto distance distribution for 100k subset.}
\begin{tabular}{c|cc|c}
\toprule
\textbf{Statistic} & \textbf{Random subset} & \textbf{Diverse subset} & \textbf{Difference} \\
\midrule
p10 & 0.881 & 0.915 & 0.034 \\
p25 & 0.901 & 0.939 & 0.038 \\
mean & 0.920 & 0.962 & 0.042 \\
median & 0.922 & 0.967 & 0.045 \\
p75 & 0.940 & 0.995 & 0.055 \\
p90 & 0.957 & 1.000 & 0.043 \\
\midrule
Wasserstein distance & \multicolumn{2}{c|}{0.042} & - \\
K-S statistic & \multicolumn{2}{c|}{0.507} & - \\
\bottomrule
\end{tabular}
\label{table_appendix_diverse_subset_100k}
\end{table}

\begin{table}[]
\centering
\setlength{\belowcaptionskip}{2pt}
\caption{Tanimoto distance distribution for 1M subset.}
\begin{tabular}{c|cc|c}
\toprule
\textbf{Statistic} & \textbf{Random subset} & \textbf{Diverse subset} & \textbf{Difference} \\
\midrule
p10 & 0.881 & 0.904 & 0.023 \\
p25 & 0.901 & 0.926 & 0.025 \\
mean & 0.920 & 0.948 & 0.028 \\
median & 0.921 & 0.950 & 0.029 \\
p75 & 0.940 & 0.974 & 0.034 \\
p90 & 0.957 & 1.000 & 0.043 \\
\midrule
Wasserstein distance & \multicolumn{2}{c|}{0.029} & - \\
K-S statistic & \multicolumn{2}{c|}{0.359} & - \\
\bottomrule
\end{tabular}
\label{table_appendix_diverse_subset_1M}
\end{table}

\begin{table}[]
\centering
\setlength{\belowcaptionskip}{2pt}
\caption{Tanimoto distance distribution for 5M subset.}
\begin{tabular}{c|cc|c}
\toprule
\textbf{Statistic} & \textbf{Random subset} & \textbf{Diverse subset} & \textbf{Difference} \\
\midrule
p10 & 0.881 & 0.896 & 0.015 \\
p25 & 0.901 & 0.917 & 0.016 \\
mean & 0.920 & 0.938 & 0.018 \\
median & 0.921 & 0.939 & 0.018 \\
p75 & 0.940 & 0.962 & 0.022 \\
p90 & 0.957 & 0.981 & 0.024 \\
\midrule
Wasserstein distance & \multicolumn{2}{c|}{0.019} & - \\
K-S statistic & \multicolumn{2}{c|}{0.241} & - \\
\bottomrule
\end{tabular}
\label{table_appendix_diverse_subset_5M}
\end{table}

\begin{table}[]
\centering
\setlength{\belowcaptionskip}{2pt}
\caption{Tanimoto distance distribution for 10M subset.}
\begin{tabular}{c|cc|c}
\toprule
\textbf{Statistic} & \textbf{Random subset} & \textbf{Diverse subset} & \textbf{Difference} \\
\midrule
p10 & 0.881 & 0.893 & 0.012 \\
p25 & 0.901 & 0.913 & 0.012 \\
mean & 0.920 & 0.934 & 0.014 \\
median & 0.921 & 0.935 & 0.014 \\
p75 & 0.940 & 0.957 & 0.017 \\
p90 & 0.957 & 0.975 & 0.018 \\
\midrule
Wasserstein distance & \multicolumn{2}{c|}{0.014} & - \\
K-S statistic & \multicolumn{2}{c|}{0.189} & - \\
\bottomrule
\end{tabular}
\label{table_appendix_diverse_subset_10M}
\end{table}

\section{Pipeline molecule filtering details}
\label{appendix_pipeline_filtering_details}

In Table \ref{appendix_dataset_workflow_statistics}, we present exact numbers for all datasets and pipeline steps, a detailed version of Table \ref{dataset_workflow_statistics} from the main body.

\begin{table}[H]
\caption{Statistics of datasets: number of molecules removed at each step and the final dataset size.}
\setlength{\tabcolsep}{6pt}
\begin{tabular}{lrrrrr}
\toprule
\textbf{Dataset} & 
\multicolumn{1}{l}{\textbf{Initial count}} & 
\multicolumn{1}{l}{\textbf{Preprocessing}} & 
\multicolumn{1}{l}{\textbf{Standardization}} & 
\multicolumn{1}{l}{\textbf{Filtering}} & 
\multicolumn{1}{l}{\textbf{Final dataset}} \\
\midrule
UniChem       & 189058653 & 0        & -458698  & -4677263 & 183922692 \\
PubChem       & 121440975 & -466660  & -107100  & -4233834 & 116633381 \\
Mcule         & 43580777  & -104840  & -157     & -13592   & 43462188  \\
ChemSpace     & 7831419   & -78      & -107     & -4345    & 7826889   \\
SuperNatural3 & 1205199   & -2913    & -331     & -44135   & 1157820   \\
COCONUT       & 695133    & -8584    & -154     & -25963   & 660432    \\
\hline
\addlinespace[2pt]
ChEMBL        & 2.4M  & -8      & -911   & -41393  & 2.4M \\
GDB-17        & 50M & -4267  & -5     & 0        & 50M \\
ZINC          & 13.7M & -944883 & -8743 & -3026   & 12.7M \\
\hline
\addlinespace[2pt]
\textbf{MolPILE}       &           &          &          &          & 221950487 \\
\bottomrule
\end{tabular}
\label{appendix_dataset_workflow_statistics}
\end{table}

\section{Full SAScore table}
\label{appendix_sascore_table}

Table \ref{appendix_sascore_table} contains distribution information for SAScore values: minimum, maximum, mean, median, Q1 and Q3, and percentiles 1, 5, 95 and 99. Those results are for filtered datasets

\begin{table}[H]
\caption{Synthesizability Score (SAScore) distributions}
\begin{tabular}{l|c|c|c|c|c|c|c|c|c|c}
\toprule
\textbf{Dataset} & \textbf{min} & \textbf{p1} & \textbf{p5} & \textbf{Q1} & \textbf{mean} & \textbf{median} & \textbf{Q3} & \textbf{p95} & \textbf{p99} & \textbf{max} \\
\hline
\addlinespace[2pt]
UniChem & 1 & 1.91 & 2.18 & 2.67 & 3.26 & 3.11 & 3.67 & 4.86 & 6.08 & 10 \\
PubChem & 1 & 1.87 & 2.14 & 2.63 & 3.25 & 3.08 & 3.66 & 4.93 & 6.22 & 10 \\
Mcule & 1 & 1.86 & 2.09 & 2.45 & 2.80 & 2.74 & 3.08 & 3.69 & 4.33 & 8.92 \\
ChemSpace & 1 & 1.80 & 2.05 & 2.47 & 2.91 & 2.82 & 3.26 & 4.06 & 4.85 & 9.47 \\
SuperNatural3 & 1 & 1.98 & 2.43 & 3.44 & 4.42 & 4.28 & 5.18 & 6.94 & 7.97 & 9.53 \\
COCONUT & 1 & 1.88 & 2.31 & 3.29 & 4.31 & 4.20 & 5.12 & 6.84 & 7.92 & 9.53 \\
\hline
\addlinespace[2pt]
\textbf{MolPILE} & 1 & 1.92 & 2.17 & 2.63 & 3.20 & 3.05 & 3.59 & 4.78 & 6 & 10 \\
\addlinespace[2pt]
\hline
\addlinespace[2pt]
ChEMBL & 1 & 1.81 & 2.08 & 2.54 & 3.18 & 2.97 & 3.58 & 5.06 & 6.50 & 9.01 \\
GDB-17 & 1 & 3.14 & 3.66 & 4.55 & 5.07 & 5.09 & 5.65 & 6.31 & 6.73 & 8.19 \\
ZINC & 1 & 1.84 & 2.11 & 2.59 & 3.08 & 3.02 & 3.47 & 4.32 & 5.10 & 7.99 \\
\bottomrule
\end{tabular}
\label{appendix_table_sascore}
\end{table}

\section{Molecular descriptors analysis}
\label{appendix_molecular_descriptors}

In this section, we analyze the distributions of molecular descriptors in MolPILE and other datasets used for model pretraining: ChEMBL, GDB-17, and ZINC. Distribution statistics are provided in Tables \ref{appendix_table_descriptors_molecular_weight}, \ref{appendix_table_descriptors_num_atoms}, \ref{appendix_table_descriptors_hba}, \ref{appendix_table_descriptors_hbd}, \ref{appendix_table_descriptors_logp}, \ref{appendix_table_tpsa2} and \ref{appendix_table_descriptors_rotb}. Values for ChEMBL, GDB-17 and ZINC are provided without the feasibility filter to avoid skewing the results.

First, ChEMBL clearly contains some outliers and unreasonable molecules, as evidences by maximum value in each descriptor, e.g. maximum molecular weight about 12.5 thousand daltons or 360 rotatable bonds. GDB-17 shows very low numbers in all regards, which is a consequence of its combinatorial construction and using at most 17 atoms. ZINC shows very low diversity and contains a very conservative set of typical medicinal compounds. Its highest molecular weight is under 1000 daltons, and the largest number of atoms is 60, not covering e.g. many oncological drugs. Similarly, there are at most 20 HBA, 15 HBD, and 45 rotatable bonds, indicating a very conservative distribution in terms of structure flexibility. We note that this may also be due to our choice of a representative subset of ZINC with established 3D structures, but the general trend points to low diversity of that dataset in terms of physicochemical properties.

\begin{table}[H]
\centering
\caption{Molecular weight distribution statistics.}
\resizebox{\textwidth}{!}{
\begin{tabular}{|l|r|r|r|r|r|r|r|r|r|r|}
\hline
\textbf{Dataset} & \textbf{min} & \textbf{p1} & \textbf{p5} & \textbf{Q1} & \textbf{mean} & \textbf{median} & \textbf{Q3} & \textbf{p95} & \textbf{p99} & \textbf{max} \\ \hline
ChEMBL & 4.0 & 181.2 & 240.3 & 325.4 & 436.4 & 393.5 & 476.4 & 707.2 & 1457.8 & 12546.3 \\ \hline
GDB-17 & 30.1 & 192.3 & 207.2 & 224.3 & 236.3 & 237.3 & 242.3 & 262.3 & 320.2 & 703.8 \\ \hline
ZINC & 51.1 & 170.2 & 237.2 & 319.4 & 383.7 & 374.5 & 440.9 & 553.4 & 655.9 & 995.1 \\ \hline
MolPILE & 1.00 & 156.20 & 210.23 & 297.31 & 383.53 & 364.39 & 436.56 & 629.64 & 923.98 & 2500.00 \\ \hline
\end{tabular}
}
\label{appendix_table_descriptors_molecular_weight}
\end{table}

\begin{table}[H]
\centering
\caption{Number of atoms distribution statistics.}
\resizebox{\textwidth}{!}{
\begin{tabular}{|l|r|r|r|r|r|r|r|r|r|r|}
\hline
\textbf{Dataset} & \textbf{min} & \textbf{p1} & \textbf{p5} & \textbf{Q1} & \textbf{mean} & \textbf{median} & \textbf{Q3} & \textbf{p95} & \textbf{p99} & \textbf{max} \\ \hline
ChEMBL & 1.0 & 12.0 & 17.0 & 23.0 & 30.6 & 28.0 & 34.0 & 50.0 & 102.0 & 780.0 \\ \hline
GDB-17 & 2.0 & 14.0 & 15.0 & 16.0 & 16.5 & 17.0 & 17.0 & 17.0 & 17.0 & 18.0 \\ \hline
ZINC & 4.0 & 12.0 & 16.0 & 22.0 & 26.6 & 26.0 & 31.0 & 38.0 & 44.0 & 60.0 \\ \hline
MolPILE & 1.00 & 10.00 & 14.00 & 20.00 & 26.71 & 25.00 & 31.00 & 44.00 & 66.00 & 150.00 \\ \hline
\end{tabular}
}
\label{appendix_table_descriptors_num_atoms}
\end{table}

\begin{table}[H]
\centering
\caption{Hydrogen bond acceptors (HBA) distribution statistics.}
\begin{tabular}{|l|r|r|r|r|r|r|r|r|r|r|}
\hline
\textbf{Dataset} & \textbf{min} & \textbf{p1} & \textbf{p5} & \textbf{Q1} & \textbf{mean} & \textbf{median} & \textbf{Q3} & \textbf{p95} & \textbf{p99} & \textbf{max} \\ \hline
ChEMBL & 0.0 & 1.0 & 2.0 & 4.0 & 5.7 & 5.0 & 7.0 & 10.0 & 21.0 & 290.0 \\ \hline
GDB-17 & 0.0 & 1.0 & 2.0 & 3.0 & 4.3 & 4.0 & 5.0 & 6.0 & 7.0 & 12.0 \\ \hline
ZINC & 0.0 & 1.0 & 2.0 & 3.0 & 4.7 & 5.0 & 6.0 & 8.0 & 10.0 & 20.0 \\ \hline
MolPILE & 0.00 & 1.00 & 2.00 & 3.00 & 4.55 & 4.00 & 6.00 & 8.00 & 12.00 & 20.00 \\ \hline
\end{tabular}
\label{appendix_table_descriptors_hba}
\end{table}

\begin{table}[H]
\centering
\caption{Hydrogen bond donors (HBD) distribution statistics.}
\begin{tabular}{|l|r|r|r|r|r|r|r|r|r|r|}
\hline
\textbf{Dataset} & \textbf{min} & \textbf{p1} & \textbf{p5} & \textbf{Q1} & \textbf{mean} & \textbf{median} & \textbf{Q3} & \textbf{p95} & \textbf{p99} & \textbf{max} \\ \hline
ChEMBL & 0.0 & 0.0 & 0.0 & 1.0 & 2.1 & 1.0 & 2.0 & 5.0 & 19.0 & 121.0 \\ \hline
GDB-17 & 0.0 & 0.0 & 0.0 & 1.0 & 2.2 & 2.0 & 3.0 & 5.0 & 6.0 & 10.0 \\ \hline
ZINC & 0.0 & 0.0 & 0.0 & 0.0 & 1.1 & 1.0 & 2.0 & 3.0 & 4.0 & 15.0 \\ \hline
MolPILE & 0.00 & 0.00 & 0.00 & 1.00 & 1.37 & 1.00 & 2.00 & 3.00 & 6.00 & 15.00 \\ \hline
\end{tabular}
\label{appendix_table_descriptors_hbd}
\end{table}

\begin{table}[H]
\centering
\caption{Water–octanol partition coefficient logarithm (logP) distribution statistics.}
\begin{tabular}{|l|r|r|r|r|r|r|r|r|r|r|}
\hline
\textbf{Dataset} & \textbf{min} & \textbf{p1} & \textbf{p5} & \textbf{Q1} & \textbf{mean} & \textbf{median} & \textbf{Q3} & \textbf{p95} & \textbf{p99} & \textbf{max} \\ \hline
ChEMBL & -247.5 & -1.4 & 0.7 & 2.6 & 3.8 & 3.7 & 4.9 & 7.0 & 9.9 & 180.5 \\ \hline
GDB-17 & -5.2 & -2.0 & -1.3 & -0.2 & 0.7 & 0.6 & 1.5 & 2.7 & 3.5 & 7.2 \\ \hline
ZINC & -10.0 & -0.9 & 0.4 & 2.2 & 3.4 & 3.5 & 4.6 & 6.4 & 7.8 & 17.7 \\ \hline
MolPILE & -10.00 & -0.83 & 0.66 & 2.36 & 3.73 & 3.48 & 4.68 & 7.31 & 14.02 & 25.00 \\ \hline
\end{tabular}
\label{appendix_table_descriptors_logp}
\end{table}

\begin{table}[H]
\centering
\caption{Topological polar surface area (TPSA) distribution statistics.}
\resizebox{\textwidth}{!}{
\begin{tabular}{|l|r|r|r|r|r|r|r|r|r|r|}
\hline
\textbf{Dataset} & \textbf{min} & \textbf{p1} & \textbf{p5} & \textbf{Q1} & \textbf{mean} & \textbf{median} & \textbf{Q3} & \textbf{p95} & \textbf{p99} & \textbf{max} \\ \hline
ChEMBL & 0.0 & 12.9 & 30.0 & 56.9 & 97.0 & 78.3 & 104.8 & 188.7 & 552.0 & 4530.8 \\ \hline
GDB-17 & 0.0 & 15.0 & 29.5 & 55.4 & 74.5 & 73.8 & 93.2 & 121.4 & 141.0 & 216.4 \\ \hline
ZINC & 0.0 & 18.5 & 32.7 & 54.7 & 73.1 & 71.4 & 89.6 & 118.0 & 145.2 & 440.0 \\ \hline
MolPILE & 0.00 & 3.24 & 21.26 & 50.16 & 73.67 & 69.67 & 90.73 & 136.42 & 211.89 & 499.99 \\ \hline
\end{tabular}
}
\label{appendix_table_tpsa2}
\end{table}

\begin{table}[H]
\centering
\caption{Number of rotatable bonds distribution statistics.}
\begin{tabular}{|l|r|r|r|r|r|r|r|r|r|r|}
\hline
\textbf{Dataset} & \textbf{min} & \textbf{p1} & \textbf{p5} & \textbf{Q1} & \textbf{mean} & \textbf{median} & \textbf{Q3} & \textbf{p95} & \textbf{p99} & \textbf{max} \\ \hline
ChEMBL & 0.0 & 0.0 & 1.0 & 3.0 & 6.5 & 5.0 & 7.0 & 14.0 & 38.0 & 360.0 \\ \hline
GDB-17 & 0.0 & 0.0 & 0.0 & 1.0 & 2.3 & 2.0 & 4.0 & 6.0 & 8.0 & 13.0 \\ \hline
ZINC & 0.0 & 1.0 & 2.0 & 4.0 & 5.3 & 5.0 & 7.0 & 10.0 & 13.0 & 45.0 \\ \hline
MolPILE & 0.00 & 0.00 & 2.00 & 4.00 & 5.85 & 5.00 & 7.00 & 12.00 & 25.00 & 60.00 \\ \hline
\end{tabular}
\label{appendix_table_descriptors_rotb}
\end{table}

\section{Molecular filters analysis}
\label{appendix_molecular_filters}

\begin{figure}[H]
    \centering
    \includegraphics[width=1.\textwidth]{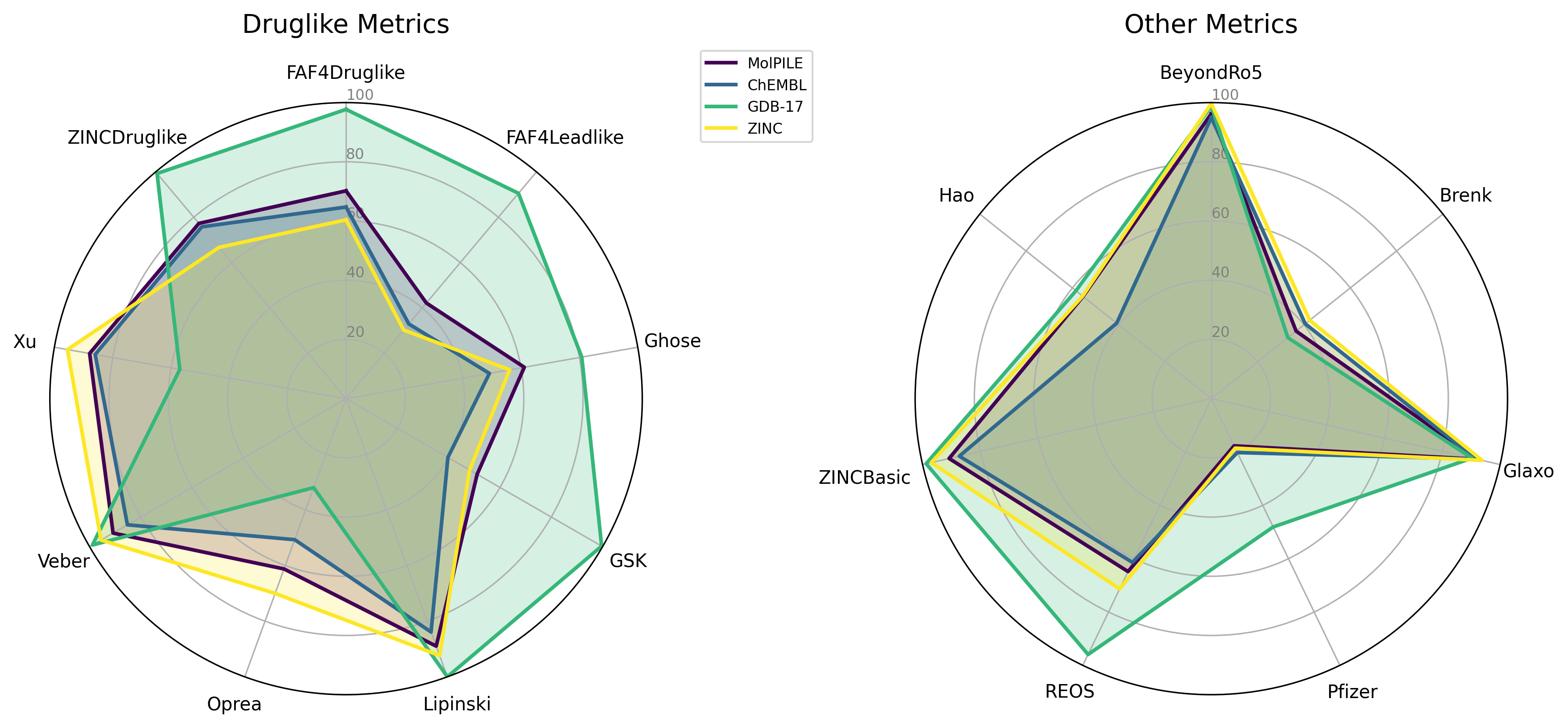}
    \caption{Percentage of molecules that fits to various molecular filters. }
    \label{appendix_figure_molecular_filters}
\end{figure}

Figure \ref{appendix_figure_molecular_filters} presents the results of molecular filter analyses. Molecular filters are used in drug design for the initial selection of molecules based on their physicochemical and structural properties. They enable the rejection of compounds with unfavorable characteristics, such as low bioavailability, difficult synthesis or the presence of reactive fragments. This narrows down the chemical space to more promising candidates and increases the efficiency of further stages of drug design. However, they also limit the chemical space, which can decrease the innovation in terms of explored compound structures.

For readability, we restrict this comparison to MolPILE, ChEMBL, GDB-17, and ZINC across 16 filters (see below for their descriptions). Molecular filters can be broadly grouped into two categories: (i) drug-likeness filters, based on structural properties of typical medicinal chemistry compounds, and (ii) filters targeting other aspects such as toxicity, reactivity, or pesticide-likeness. Examining the fraction of molecules passing each filter provides complementary insight into both physicochemical properties and characteristic substructures of the datasets \citep{ApisTox_ML}. Ideally, most molecules should pass standard filters, indicating consistency with well-established distributions. However, we also expect a subset to fall outside these boundaries, since advances in synthesis methods enable exploration of chemical space beyond conservative, rule-based definitions. In the case of physicochemical filters, this space corresponds directly to a descriptor-defined hypercube (e.g., molecular weight, logP), where strict adherence would imply limited diversity.

As expected, all datasets score highly on the Beyond Rule of 5 filter, which is designed to exclude only extreme, unreasonable molecules. This also supports the validity of our molecular feasibility filter. MolPILE shows balanced results overall, with the highest values on general filters such as Xu or Veber.

In contrast, GDB-17 fails to generalize beyond traditional chemical space, with near-complete compliance (about 100\%) across multiple filters, including Lipinski, GSK, FAF4-druglike, ZINC-druglike, Veber, ZINC-basic, REOS, and Beyond Rule of 5. Other datasets display much lower pass rates, particularly for more restrictive filters such as GSK, Ghose, FAF4-druglike, FAF4-leadlike, REOS, and ZINC-druglike.

Molecular filters used are:

\begin{enumerate}
\item \textbf{FAFDrugs4 drug-like filter} \citep{FAF4} - designed to keep only drug-like molecules, based on literature describing physico-chemical properties of drugs and their statistical analysis. Selected so that up to 90\% of the 916 FDA-approved oral drugs fulfill the rules of this filter. Molecules must fulfill conditions: molecular weight in range $[100, 600]$, logP in range $[-3, 6]$, HBA $\leq$ 12, HBD $\leq 7$, TPSA $\leq 180$, number of rotatable bonds $\leq 11$, number of rigid bonds $\leq 30$, number of rings $\leq 6$, max ring size $\leq 18$, number of carbons in range $[3, 35]$, number of heteroatoms in range $[1, 15]$, non-carbons to carbons ratio in range $[0.1, 1.1]$, number of charged functional groups $\leq 4$, total formal charge in range $[-4, 4]$.

\item \textbf{ZINC drug-like} \citep{ZINC_druglike} - designed to keep only drug-like molecules. Based only on physico-chemical properties, since SMARTS for additional rules are not publicly available. Molecules must fulfill conditions: molecular weight in range $[60, 600]$, logP in range $[-4, 6]$, HBA $\leq 11$, HBD $\leq 6$, TPSA $\leq 150$, number of rotatable bonds $\leq 12$, number of rigid bonds $\leq 50$, number of rings $\leq 7$, max ring size $\leq 12$, number of carbons $\geq 3$, non-carbons to carbons ratio $\leq 2.0$, number of charged functional groups $\leq 4$, total formal charge in range $[-4, 4]$.

\item \textbf{Rule of Xu} \citep{Xu} - another filter designed to identify drug-like molecules. Molecules must fulfill conditions: HBD $\leq 5$, HBA $\leq 10$, number of rotatable bonds in range $[2, 35]$, number of rings in range $[1, 7]$, number of heavy atoms in range $[10, 50]$.

\item \textbf{Rule of Veber} \citep{filters_Veber} - designed to identify molecules likely to exhibit good oral bioavailability. Molecules must fulfill conditions: number of rotatable bonds $\leq 10$, TPSA $\leq 140$.

\item \textbf{Oprea filter} \citep{Oprea} - filter for drug likeness, designed by comparing drug and non-drug compounds across multiple datasets. Molecules must fulfill conditions: HBD $\leq 2$, HBA in range $[2, 9]$, number of rotatable bonds in range $[2, 8]$, number of rings in range $[1, 4]$.

\item \textbf{Lipinski’s Rule of 5 (RO5)} \citep{Lipinski} - evaluates the drug-likeness of a molecule as an orally active drug, assuming that they are small and lipophilic. Also known as Pfizer's rule of 5. Molecules are allowed to violate at most one of the rules. Conditions: molecular weight $\leq 500$, HBA $\leq 10$, HBD $\leq 5$, logP $\leq 5$. Hydrogen bond acceptors (HBA) and donors (HBD) use a simplified definition, taking into consideration only oxygen and nitrogen bonds with hydrogen (OH, NH).

\item \textbf{GSK rule (4/400)} \citep{filters_GSK} - interpretable ADMET rule of thumb for drug-likeness. Molecules must fulfill conditions: molecular weight $\leq 400$, logP $\leq 4$ .

\item \textbf{Ghose filter} \citep{Ghose} - used for searching for drug-like compounds. Molecules must fulfill conditions: molecular weight in range $[160, 400]$, logP in range $[-0.4, 5.6]$, number of atoms in range $[20, 70]$, molar refractivity in range $[40, 130]$ .

\item \textbf{FAFDrugs4 lead-like filter} \citep{FAFDruglike} - based on literature describing physico-chemical properties of lead drugs, designed to find lead-like candidates, i.e. starting point molecules that can be further optimized. They should be relatively small, with low logP, and can be ``decorated'' further to increase affinity and/or selectivity, without becoming very ADMET-unfriendly. Basically a more restrictive variant of FAFDrugs4 drug-like filter. Molecules must fulfill conditions: molecular weight in range $[150, 400]$, logP in range $[-3, 4]$, HBA $\leq 7$, HBD $\leq 4$, TPSA $\leq 160$, number of rotatable bonds $\leq 9$, number of rigid bonds $\leq 30$, number of rings $\leq 4$, max ring size $\leq 18$, number of carbons in range $[3, 35]$, number of heteroatoms in range $[1, 15]$, non-carbons to carbons ratio in range $[0.1, 1.1]$, number of charged functional groups $\leq 4$, total formal charge in range $[-4, 4]$, number of stereocenters $\leq 2$.

\item \textbf{Beyond rule of 5} \citep{BeyondRO5} - designed to cover cover novel orally bioavailable drugs that do not fulfill the original conditions of Lipinski's rule of 5. Allows less typical molecules, particularly suitable for difficult targets, allowing greater flexibility. Molecules must fulfill conditions: molecular weight $\leq 1000$, logP in range $[-2, 10]$, HBA $\leq 15$, HBD $\leq 6$, TPSA $\leq 250$, number of rotatable bonds $\leq 20$ .

\item \textbf{Hao rule for pesticides} \citep{Hao} - designed to describe physicochemical properties of pesticides, for use in general pesticide design. Molecules must fulfill conditions: molecular weight $\leq 435$, logP $\leq 6$, HBD $\leq 2$, HBA $\leq 6$, number of rotatable bonds $\leq 9$, number of aromatic bonds $\leq 17$ .

\item \textbf{ZINC basic filter} \citep{ZINCBasic} - designed to keep only drug-like molecules, removing molecules with unwanted functional groups. Used by docking.org for ZINC database as basic set of filters, applied to all vendor catalogs. Substructural filter, with rules available at \cite{ZINCBasic}.

\item \textbf{REOS filter} \citep{REOS} - Rapid Elimination Of Swill (REOS) filter is designed to remove molecules with undesirable properties for general drug discovery. Molecules must fulfill conditions: molecular weight in range $[200, 500]$, logP in range $[-5, 5]$, HBA $\leq 10$, HBD $\leq 5$, charge in range $[-2, 2]$, number of rotatable bonds $\leq 8$, number of heavy atoms in range $[15, 50]$.

\item \textbf{Pfizer 3/75 rule} \citep{Pfizer_filter_1,Pfizer_filter_2} - based on observation that compounds exhibiting low partition coefficient (logP) and high topological polar surface area (TPSA) are roughly 2.5 times more likely to be free of toxicity issues in the tested conditions. Molecules must fulfill conditions: logP $\leq 3$, TPSA $\geq 75$.

\item \textbf{Glaxo filter} \citep{Glaxo} - designed to filter out molecules with reactive functional groups, unsuitable leads (i.e., compounds which would not be initially followed up), and unsuitable natural products (i.e., derivatives of natural product compounds known to interfere with common assay procedures). Rule definitions are available in the supplementary material of the original publication \citep{Glaxo}.

\item \textbf{Brenk filter} \citep{Brenk} - designed to filter out molecules containing substructures with undesirable pharmacokinetics or toxicity, e.g., sulfates, phosphates, nitro groups. Resulting set should be reasonable lead-like molecules for optimization campaigns and HTS. Rule definitions are available in the supplementary material of the original publication \citep{Brenk}.
\end{enumerate}

\section{Functional groups, scaffolds, and salts analyses}
\label{appendix_functional_groups_scaffolds_salts}

\begin{table}[h]
\centering
\caption{Number of unique scaffolds, functional groups, and salts in each dataset.}
\begin{tabular}{l r r r}
\hline
\addlinespace[1pt]
\textbf{Dataset} & \textbf{Scaffolds} & \textbf{Functional groups} & \textbf{Salts} \\
\addlinespace[1pt]
\hline
\addlinespace[1pt]
UniChem        & 3,213,417 & 122,050 &   953,082 \\
PubChem        & 3,046,726 &  79,103 &   784,495 \\
Mcule          &   310,894 &   4,317 &    28,888 \\
ChemSpace      &   259,597 &   3,140 &     9,841 \\
SuperNatural3  &    47,484 &   2,498 &       233 \\
COCONUT        &    45,268 &   1,595 &     2,002 \\
\hline
\addlinespace[1pt]
\textbf{MolPILE}        & 3,620,809 & 128,347 & 1,089,501 \\
\addlinespace[1pt]
\hline
\addlinespace[1pt]
ChEMBL         &   172,606 &   3,828 &    22,474 \\
GDB-17         &    70,377 &  24,274 &         0 \\
ZINC           &   191,514 &   5,813 &         9 \\
\hline
\end{tabular}
\label{appendix_table_functional_groups_scaffolds_salts}
\end{table}

Here, we present an analysis of Bemis-Murcko scaffolds \citep{scaffolds}, functional groups computed with Ertl's algorithm \citep{functional_groups}, and salts. Those are structural measures of diversity, focusing on different aspects of molecular graph shape. Results are summarized in Table \ref{appendix_table_functional_groups_scaffolds_salts}.

Number of unique Bemis-Murcko scaffolds is a commonly used diversity measure. Here, we count so-called ``generic scaffolds'', as described in the original paper \citep{scaffolds}, consisting of connected rings backbone, and replacing all atoms by carbons. This ensures that we focus solely on the true core of the molecule. MolPILE has the largest number of scaffolds, over 1 million, showing its very high diversity in this regard. The absolute number here is important, as this means that models trained on MolPILE will be exposed to a wide collection of molecular shapes.

For functional groups analysis, we used Ertl's algorithm, which extracts them algorithmically. There is no one commonly used list of functional groups, and their definitions vary between chemists, so relying on an algorithmic approach is more objective in this regard. We select only functional groups between 2 and 20 atoms, and only those appearing in at least 10 molecules, in order to remove computational artifacts, similar to analysis of ChEMBL in the original paper \citep{functional_groups}. Again, MolPILE contains the largest number of functional groups, which means that models pretrained on it will be exposed to various substituents. This is particularly relevant to SMILES-based transformers, as this will also impact the tokenization.

Lastly, we checked the number of salts, i.e., molecules with at least two charged disconnected components (in terms of covalent bonds). This is very important for SMILES-based transformers, as without this they won't even recognize the fact of a molecule having multiple fragments. The SMILES format uses the dot symbol ``.'' to mark disconnected components, and it must be a part of the tokenizer and have a reasonable learned embedding in order to be useful. Interestingly, GDB-17 does not contain any salts at all, and ZINC only trace number of 9. This means that models trained on those datasets won't work on salts at all, while this is highly relevant to agrochemistry \citep{QSAR_workflows,ApisTox}. Such models may even error out on such SMILES strings, depending on the implementation. MolPILE contains over 1 million salts, about $0.5\%$, which exposes all models to a reasonable number of them, and allows proper learning.

\section{Evaluation datasets and model training}
\label{appendix_model_training_evaluation_datasets}

Here, we describe evaluation details and model training procedure in detail. We also provide results on individual datasets.

\subsection{Datasets and benchmarks}

To evaluate models retrained on MolPILE, we use a total of 52 datasets from 2 general benchmarks for drug design and ADMET (MoleculeNet, TDC), ligand-based virtual screening benchmark (WelQrate), and two datasets from two distinctly different areas: protein-ligand binding (ToxBench), and ecotoxicology (ApisTox).

\textbf{MoleculeNet} \citep{MoleculeNet} is the most commonly used benchmark for evaluating ML models for molecular property prediction. It contains 16 datasets, 9 for classification and 7 for regression. However, in literature most commonly used are 8 classification datasets and 3 regression datasets, excluding PCBA (due to extreme size), PDBBind (due to relatively low quality as a protein-ligand binding dataset), and QM7, QM8 and QM9 (quantum chemistry, requiring dedicated models and 3D information). Thus, we use the following 11 datasets: BACE, BBBP, HIV (single task classification), ClinTox, MUV, SIDER, Tox21, ToxCast (multi-task classification), ESOL, FreeSolv, and Lipophilicity (regression). AUROC is used for all classification datasets, and MAE for all regression ones.

\textbf{Therapeutics Data Commons (TDC} \citep{TDC} is a large collection of diverse tasks focusing on medicinal chemistry, covering a wide variety of ADMET tasks. We select all datasets between around 500 and 50000 molecules, to have reasonable data size for training and testing. This amounts to 19 datasets. See tables below for all dataset names. AUROC is used for all classification datasets, and MAE for all regression ones.

\textbf{WelQrate} \citep{WelQrate} has been designed as a ``gold standard'' benchmark for ligand-based virtual screening. It consists of 9 large-scale datasets from 5 therapeutic target classes, which underwent expert curation and rigorous process of categorization, verification, filtering and cleaning. It evaluates the ability of models to not only predict bioactivity and protein-ligand, based only on the ligand structure, but also to rank them from most to least promising. This is measured by using the BEDROC metric (Boltzmann-enhanced discrimination of receiver operating characteristic) \citep{BEDROC}.

\textbf{ToxBench} \citep{ToxBench} is a dataset based on AB-FEP calculations, covering 8770 ligand–ER$\alpha$ complexes with calculated binding free energies, some of which have been experimentally verified. In particular, it features carefully prepared train-test data splits, enabling a reliable assessment of models for protein-ligand binding. We use it with a ligand-only approach, similar to ChemProp in \cite{ToxBench}, embedding the ligand and predicting the binding free energy for ER$\alpha$ protein. The recommended metric, as usual in protein-ligand binding, is RMSE.

\textbf{ApisTox} \citep{ApisTox,ApisTox_ML} is a dataset for predicting pesticides toxicity to honey bees. It uses binarized LD50, following US EPA recommendation, resulting in binary classification task. We use the time split, which uses the newest pesticides as the test set, approximating the realistic conditions of novel pesticide design. Following recommendations, we use AUROC as a metric.

\subsection{Model training}

We retrain Mol2vec \citep{Mol2vec} following the original authors' implementation, based on Gensim, using CPU only. Embeddings have 300 dimensions, and are trained using Skip-gram approach for 5 epochs. For ChemBERTa \citep{ChemBERTa}, we use HuggingFace transformers and MLM modeling, exactly following the original authors' code. The only difference is that we ensure that the model sees the whole dataset at least once before early stopping, and we evaluate every 500 steps, rather than 50, as the original settings stopped training abnormally quickly.

We use the same classification head in all cases for fair comparison: Random Forest (RF) with 500 trees and entropy or MSE split function (depending on dataset). This choice was made since RF performs very well on average, does not require extensive hyperparameter tuning, and is commonly used in chemoinformatics. It also supports multitask datasets out-of-the-box in the scikit-learn implementation. We do not perform any hyperparameter tuning, and use validation data for training in datasets already pre-split into training-validation-testing subsets (e.g. MoleculeNet, TDC).

\subsection{Detailed results}

Tables below contain detailed results of original and retrained models for Mol2vec and ChemBERTa on all datasets.

\begin{table}[ht]
\centering
\caption{Mol2vec results comparison on MoleculeNet classification tasks.}
\begin{tabular}{lccc}
\toprule
\textbf{Dataset} & \textbf{AUROC original} & \textbf{AUROC retrained} & \textbf{Gain} \\
\midrule
BACE     & 81.07 & 78.86 & -2.21 \\
BBBP     & 72.46 & 71.96 & -0.50 \\
HIV      & 77.46 & 77.37 & -0.09 \\
ClinTox  & 66.27 & 80.31 & +14.04 \\
MUV      & 66.10 & 70.10 & +4.00 \\
SIDER    & 67.61 & 66.79 & -0.82 \\
Tox21    & 77.49 & 76.82 & -0.67 \\
ToxCast  & 65.27 & 65.83 & +0.56 \\
\midrule
\textbf{Average} & 71.72 & 73.51 & \textbf{+1.79} \\
\bottomrule
\end{tabular}
\end{table}

\begin{table}[ht]
\centering
\caption{Mol2vec comparison on MoleculeNet regression tasks.}
\begin{tabular}{lccc}
\toprule
\textbf{Dataset} & \textbf{MAE original} & \textbf{MAE retrained} & \textbf{Gain} \\
\midrule
ESOL            & 0.846 & 0.798 & -0.048\\
FreeSolv        & 2.876 & 2.327 & -0.549\\
Lipophilicity   & 0.709 & 0.697 & -0.012\\
\midrule
\textbf{Average} & 1.477 & 1.274 & \textbf{-0.203} \\
\bottomrule
\end{tabular}
\end{table}

\begin{table}[ht]
\centering
\caption{Mol2vec results comparison on TDC classification tasks.}
\begin{tabular}{lccc}
\toprule
\textbf{Dataset} & \textbf{AUROC original} & \textbf{AUROC retrained} & \textbf{Gain} \\
\midrule
ames                              & 80.57 & 81.06 & +0.49 \\
bioavailability\_ma               & 73.26 & 71.22 & -2.04 \\
cyp1a2\_veith                     & 91.37 & 91.92 & +0.55 \\
cyp2c9\_veith                     & 86.18 & 86.71 & +0.53 \\
cyp2c9\_substrate\_carbonmangels  & 61.79 & 62.64 & +0.85 \\
cyp2c19\_veith                    & 84.88 & 85.53 & +0.65 \\
cyp2d6\_veith                     & 84.11 & 84.24 & +0.13 \\
cyp2d6\_substrate\_carbonmangels  & 81.60 & 82.41 & +0.81 \\
cyp3a4\_veith                     & 84.21 & 84.98 & +0.77 \\
cyp3a4\_substrate\_carbonmangels  & 65.38 & 64.72 & -0.66 \\
dili                              & 91.76 & 95.07 & +3.31 \\
herg                              & 82.03 & 83.46 & +1.43 \\
herg\_karim                       & 85.77 & 86.04 & +0.27 \\
hia\_hou                          & 97.43 & 98.93 & +1.50 \\
pampa\_ncats                      & 69.88 & 71.31 & +1.43 \\
pgp\_broccatelli                  & 89.11 & 87.57 & -1.54 \\
sarscov2\_3clpro\_diamond         & 68.92 & 70.22 & +1.30 \\
sarscov2\_vitro\_touret           & 56.14 & 59.01 & +2.87 \\
\midrule
\textbf{Average}                  & 79.69 & 80.39 & \textbf{+0.58} \\
\bottomrule
\end{tabular}
\end{table}

\begin{table}[]
\centering
\caption{Mol2vec results comparison on TDC regression tasks.}
\begin{tabular}{lccr}
\toprule
\multicolumn{1}{c}{\textbf{Dataset}} & \textbf{MAE original} & \textbf{MAE retrained} & \multicolumn{1}{c}{\textbf{Gain}} \\
\midrule
caco2\_wang & 0.303 & 0.311 & 0.008 \\
clearance\_hepatocyte\_az & 36.953 & 37.425 & 0.472 \\
clearance\_microsome\_az & 28.449 & 30.415 & 1.966 \\
half\_life\_obach & 19.444 & 13.273 & -6.171 \\
ld50\_zhu & 0.666 & 0.656 & -0.01 \\
ppbr\_az & 9.393 & 9.552 & 0.159 \\
solubility\_aqsoldb & 1.007 & 0.974 & -0.033 \\
vdss\_lombardo & 4.119 & 3.762 & -0.357 \\
\midrule
\textbf{Average} & 12.542 & 12.046 & \textbf{-0.496} \\
\bottomrule
\end{tabular}
\end{table}

\begin{table}[ht]
\centering
\caption{Mol2vec results comparison on WelQrate datasets.}
\begin{tabular}{lcccccc}
\toprule
\textbf{Dataset} & \multicolumn{2}{c}{\textbf{Original}} & \multicolumn{2}{c}{\textbf{Retrained}} & \multicolumn{2}{c}{\textbf{Gain}} \\
\cmidrule(lr){2-3} \cmidrule(lr){4-5} \cmidrule(lr){6-7}
 & AUROC & BEDROC & AUROC & BEDROC & AUROC & BEDROC \\
\midrule
AID1798    & 63.12 & 20.35 & 66.46 & 20.81 & +3.34 & +0.46 \\
AID1843    & 76.89 & 47.47 & 76.97 & 49.35 & +0.08 & +1.88 \\
AID2258    & 70.12 & 32.51 & 73.17 & 37.25 & +3.05 & +4.74 \\
AID2689    & 74.40 & 46.48 & 73.97 & 43.94 & -0.43 & -2.54 \\
AID435008  & 69.52 & 35.83 & 73.83 & 39.44 & +4.31 & +3.61 \\
AID435034  & 70.48 & 22.52 & 75.07 & 25.59 & +4.59 & +3.07 \\
AID463087  & 86.24 & 43.37 & 86.23 & 42.26 & -0.01 & -1.11 \\
AID485290  & 75.09 & 35.87 & 76.58 & 37.73 & +1.49 & +1.86 \\
AID488997  & 73.09 & 36.00 & 72.44 & 37.72 & -0.65 & +1.72 \\
\midrule
\textbf{Average} & 73.22 & 35.60 & 74.97 & 37.12 & \textbf{+1.75} & \textbf{+1.52} \\
\bottomrule
\end{tabular}
\end{table}

\begin{table}[t]
\centering
\caption{Mol2vec results comparison on ToxBench dataset.}
\begin{tabular}{lcccc}
\toprule
\multicolumn{1}{c}{\textbf{Dataset}} & \textbf{Metric} & \textbf{Original} & \textbf{Retrained} & \textbf{Gain} \\
\midrule
\multirow{2}{*}{ToxBench} & Pearson correlation & 0.535 & 0.579 & 0.044 \\
 & RMSE & 3.982 & 3.843 & -0.139 \\
\bottomrule
\end{tabular}
\end{table}

\begin{table}[t]
\centering
\caption{Mol2vec results comparison on ApisTox dataset.}
\begin{tabular}{lccc}
\toprule
\multicolumn{1}{c}{\textbf{Dataset}} & \textbf{AUROC original} & \textbf{AUROC retrained} & \textbf{Gain} \\
\midrule
ApisTox & 70.34 & 72.57 & +2.23 \\
\bottomrule
\end{tabular}
\end{table}

\begin{table}[ht]
\centering
\caption{ChemBERTa results comparison on MoleculeNet classification tasks.}
\begin{tabular}{lccc}
\toprule
\textbf{Dataset} & \textbf{AUROC original} & \textbf{AUROC retrained} & \textbf{Gain} \\
\midrule
BACE     & 71.52 & 73.80 & +2.28 \\
BBBP     & 73.93 & 71.65 & -2.28 \\
HIV      & 74.11 & 74.54 & +0.43 \\
ClinTox  & 98.36 & 99.41 & +1.05 \\
MUV      & 49.58 & 53.54 & +3.96 \\
SIDER    & 59.58 & 61.86 & +2.28 \\
Tox21    & 68.77 & 69.81 & +1.04 \\
ToxCast  & 59.20 & 59.15 & -0.05 \\
\midrule
\textbf{Average} & 69.38 & 70.47 & \textbf{+1.09} \\
\bottomrule
\end{tabular}
\end{table}

\begin{table}[ht]
\centering
\caption{ChemBERTa comparison on MoleculeNet regression tasks.}
\begin{tabular}{lccc}
\toprule
\textbf{Dataset} & \textbf{MAE original} & \textbf{MAE retrained} & \textbf{Gain} \\
\midrule
ESOL            & 1.299 & 0.998 & -0.301 \\
FreeSolv        & 2.478 & 2.169 & -0.309 \\
Lipophilicity   & 0.760 & 0.759 & -0.001 \\
\midrule
\textbf{Average} & 1.512 & 1.309 & \textbf{-0.200} \\
\bottomrule
\end{tabular}
\end{table}

\begin{table}[ht]
\centering
\caption{ChemBERTa results comparison on TDC classification tasks.}
\begin{tabular}{lccc}
\toprule
\textbf{Dataset} & \textbf{AUROC original} & \textbf{AUROC retrained} & \textbf{Gain} \\
\midrule
ames                             & 74.40 & 74.85 & +0.45 \\
bioavailability\_ma              & 67.94 & 68.17 & +0.23 \\
cyp1a2\_veith                    & 87.38 & 87.38 & +0.00 \\
cyp2c9\_veith                    & 81.75 & 83.87 & +2.12 \\
cyp2c9\_substrate\_carbonmangels & 62.38 & 63.99 & +1.61 \\
cyp2c19\_veith                   & 81.79 & 83.16 & +1.37 \\
cyp2d6\_veith                    & 77.96 & 80.64 & +2.68 \\
cyp2d6\_substrate\_carbonmangels & 74.84 & 78.46 & +3.62 \\
cyp3a4\_veith                    & 80.51 & 82.45 & +1.94 \\
cyp3a4\_substrate\_carbonmangels & 62.62 & 59.67 & -2.95 \\
dili                             & 67.13 & 75.04 & +7.91 \\
herg                             & 68.42 & 72.37 & +3.95 \\
herg\_karim                      & 79.69 & 81.82 & +2.13 \\
hia\_hou                         & 81.34 & 88.35 & +7.01 \\
pampa\_ncats                     & 64.34 & 69.29 & +4.95 \\
pgp\_broccatelli                 & 83.54 & 82.46 & -1.08 \\
sarscov2\_3clpro\_diamond        & 63.84 & 59.63 & -4.21 \\
sarscov2\_vitro\_touret          & 59.36 & 50.25 & -9.11 \\
\midrule
\textbf{Average}                 & 73.29 & 74.55 & \textbf{+1.26} \\
\bottomrule
\end{tabular}
\end{table}

\begin{table}[ht!]
\centering
\caption{ChemBERTa results comparison on TDC regression tasks.}
\begin{tabular}{lccr}
\toprule
\multicolumn{1}{c}{\textbf{Dataset}} & \textbf{MAE original} & \textbf{MAE retrained} & \multicolumn{1}{c}{\textbf{Gain}} \\
\midrule
caco2\_wang              & 0.473  & 0.444  & -0.029 \\
clearance\_hepatocyte\_az & 37.492 & 38.493 & +1.001 \\
clearance\_microsome\_az  & 30.402 & 30.289 & -0.113 \\
half\_life\_obach        & 29.757 & 16.357 & -13.400 \\
ld50\_zhu                & 0.692  & 0.686  & -0.006 \\
ppbr\_az                 & 10.446 & 10.567 & +0.121 \\
solubility\_aqsoldb      & 1.309  & 1.211  & -0.098 \\
vdss\_lombardo           & 3.524  & 4.709  & +1.185 \\
\midrule
\textbf{Average}         & 14.262 & 12.845 & \textbf{-1.417} \\
\bottomrule
\end{tabular}
\end{table}

\begin{table}[ht!]
\centering
\caption{ChemBERTa results comparison on WelQrate datasets.}
\begin{tabular}{lcccccc}
\toprule
\textbf{Dataset} & \multicolumn{2}{c}{\textbf{Original}} & \multicolumn{2}{c}{\textbf{Retrained}} & \multicolumn{2}{c}{\textbf{Gain}} \\
\cmidrule(lr){2-3} \cmidrule(lr){4-5} \cmidrule(lr){6-7}
 & AUROC & BEDROC & AUROC & BEDROC & AUROC & BEDROC \\
\midrule
AID1798    & 60.71 & 15.63 & 64.58 & 18.67 & +3.87 & +3.04 \\
AID1843    & 66.12 & 30.80 & 64.39 & 29.55 & -1.73 & -1.25 \\
AID2258    & 66.29 & 30.02 & 68.13 & 30.70 & +1.84 & +0.68 \\
AID2689    & 58.47 & 24.61 & 65.46 & 32.31 & +6.99 & +7.70 \\
AID435008  & 63.60 & 29.10 & 61.43 & 25.45 & -2.17 & -3.65 \\
AID435034  & 62.41 & 25.16 & 63.57 & 24.59 & +1.16 & -0.57 \\
AID463087  & 73.12 & 31.01 & 76.88 & 33.63 & +3.76 & +2.62 \\
AID485290  & 67.58 & 25.64 & 66.22 & 24.51 & -1.36 & -1.13 \\
AID488997  & 61.17 & 23.16 & 61.86 & 25.31 & +0.69 & +2.15 \\
\midrule
\textbf{Average} & 64.39 & 26.13 & 65.99 & 28.26 & \textbf{+1.45} & \textbf{+1.07} \\
\bottomrule
\end{tabular}
\end{table}

\begin{table}[ht!]
\centering
\caption{ChemBERTa results comparison on ToxBench dataset.}
\begin{tabular}{lcccc}
\toprule
\multicolumn{1}{c}{\textbf{Dataset}} & \textbf{Metric} & \textbf{Original} & \textbf{Retrained} & \textbf{Gain} \\
\midrule
\multirow{2}{*}{ToxBench} 
 & Pearson correlation & 0.448 & 0.447 & -0.001 \\
 & RMSE                & 4.241 & 4.224 & -0.017 \\
\bottomrule
\end{tabular}
\end{table}

\begin{table}[ht!]
\centering
\caption{ChemBERTa results comparison on ApisTox dataset.}
\begin{tabular}{lccc}
\toprule
\multicolumn{1}{c}{\textbf{Dataset}} & \textbf{AUROC original} & \textbf{AUROC retrained} & \textbf{Gain} \\
\midrule
ApisTox & 71.65 & 68.84 & -2.81 \\
\bottomrule
\end{tabular}
\end{table}







\end{document}